%% file: main.tex
\documentclass[10pt,twocolumn,letterpaper]{article}

\usepackage[]{cvpr}      


\definecolor{cvprblue}{rgb}{0.21,0.49,0.74}
\usepackage[pagebackref,breaklinks,colorlinks,allcolors=cvprblue]{hyperref}
\usepackage{graphicx,subcaption, float}

\usepackage{xcolor}
\usepackage[table]{xcolor}
\usepackage{hyperref}
\hypersetup{
    colorlinks=true,
    linkcolor=magenta,  
    citecolor=magenta,
    urlcolor=magenta
}

\usepackage{url}
\usepackage{xcolor}
\definecolor{newcolor}{rgb}{.8,.349,.1}

\usepackage{algorithm}
\usepackage{algpseudocode}
\usepackage{subcaption}
\usepackage{multirow}
\usepackage{xcolor}
\usepackage{graphics}
\usepackage{booktabs}
\usepackage{bbding}

\title{A Parameter-Efficient Three-Branch Architecture for Multimodal Misinformation Detection with Limited Annotations}

\author{
\textbf{Daniele Cardullo$^1$, Simone Teglia$^1$, Irene Amerini$^1$} \\
$^1$Sapienza University of Rome \\
\textit{cardullo.2127806@studenti.uniroma1.it}\\ 
\textit{\{teglia, amerini\}@diag.uniroma1.it}
}

\begin{document}
\twocolumn[{%
\renewcommand\twocolumn[1][]{#1}%
\maketitle
\begin{center}
    \captionsetup{type=figure}
\end{center}%
}]
\input{sec/0_abstract}    
\input{sec/1_intro}
\input{sec/2_related_work}
\input{sec/3_method}
\input{sec/4_experiments}
\input{sec/5_results}
\input{sec/6_conclusions}
{
    \small
    \bibliographystyle{ieeenat_fullname}
    \bibliography{main}
}


\end{document}

%% file: sec/0_abstract.tex
\begin{abstract}
With the rise of easily accessible generative tools for creating and manipulating multimedia content, the threat of realistic synthetic alterations to digital media, often involving manipulations across multiple modalities simultaneously, has grown exponentially. Recently, such techniques have been increasingly employed to distort narratives of important events and to spread misinformation on social media, prompting the development of misinformation detectors. In the context of misinformation conveyed through image-text pairs, several detection methods have been proposed. However, these approaches typically rely on computationally intensive architectures that often require large amounts of annotated data.
In this work we introduce \textbf{LADLE-MM}: \textbf{L}imited \textbf{A}nnotation based \textbf{D}etector with \textbf{L}earned \textbf{E}nsembles for \textbf{M}ultimodal \textbf{M}isinformation, a model-soup initialized multimodal misinformation detector designed to operate under limited annotation setup and constrained training resources. LADLE-MM is composed of two unimodal branches and a third multimodal one that enhances image and text representations with additional multimodal embeddings extracted from BLIP, serving as fixed reference space. 
Tested on the DGM4 benchmark, our approach obtains competitive performance on both binary and multi-label classification tasks. Despite using 60.3\% fewer trainable parameters than previous state-of-the-art models, it outperforms existing methods when trained under a limited annotation setup, specifically without grounding annotations.
Moreover, when evaluated on the VERITE dataset, LADLE-MM outperforms current state-of-the-art-approaches that utilize more complex architectures involving Large Vision-Language-Models, demonstrating the effective generalization ability in an open-set setting and strong robustness to unimodal bias. We publicly release our code at \url{https://github.com/ALCOR-Lab-DIAG/LADLE-MM}.
\\
\end{abstract}

%% file: sec/1_intro.tex
\section{Introduction}
\label{sec:intro}
In recent years, generative artificial intelligence has gained increasing attention due to its rapid progress and wide range of applications. Generative models are capable of producing photorealistic images and generating human-like textual content, making it increasingly difficult to distinguish between authentic and synthetic information. The introduction of models such as Variational Autoencoders (VAEs) \cite{vae-intro-kingma2019}, Generative Adversarial Networks (GANs) \cite{gan-goodfellow2014}, and more recently, Diffusion Models (DMs) \cite{diffusion-ldm-rombach2021}, has enabled the generation of highly realistic visual content and facilitated the manipulation of images with relatively low effort. In the area of Natural Language Processing (NLP), Large Language Models (LLMs) like GPT \cite{openai2023gpt4}, Claude \cite{TheC3} and Gemini \cite{gemini} have significantly improved the ability to generate coherent and diverse textual outputs, ranging from brief messages to entire books \cite{llm-textgen-overview-2024}, contributing to major advancements in fields  like Natural Language Understanding (NLU) and Text Generation.  The integration of vision and language models has led to the development of multimodal generative systems capable of producing images from textual prompts \cite{diffusion-ldm-rombach2021, t2i-dalle-ramesh2021} or produce descriptive captions from visual inputs \cite{captioning-show-and-tell-vinyals2015, captioning-fast-multi-length-2023}. These systems demonstrate a growing capacity to process and synthesize information across different modalities. At the same time, the increasing accessibility of these generative tools has raised serious concerns regarding their potential misuse. The synthetic production of images and texts can support misinformation, identity fraud, and social engineering attacks \cite{fraud-ai-images-hks-review-2024, deepfake-criminal-justice-review-2024, deepfake-news-entertainment-aisoc-2025}, posing new challenges for digital authenticity and trust.
The combined use of generative models for images and texts to alter and manipulate information raises the problem of \textit{multimodal misinformation}. This is a potentially disruptive issue that can cause both social and economic harm. To combat the rise of such threat and to improve digital trust it is necessary to develop accurate yet cost-effective solutions for misinformation detection.

First unimodal approaches to misinformation detection either processed text or images alone \cite{xiao2024msynfd,cao2020exploring,ghai2024deep}, failing to account for the rich and cross-modal nature of modern online multimedia content. 
In contrast to unimodal methods, multimodal misinformation detection models often leverage contrastive learning \cite{mm-fakenews-contrastive-wang2023}, cross-modal attention mechanisms \cite{dgm4}, or fusion-based architectures \cite{mm-deepfake-detection-salvi2023} to capture inconsistencies between visual and textual modalities. These approaches have shown promising results in identifying subtle alterations in visual and text modalities but they still face some limitations. However, current state-of-the-art models heavily rely on large-scale annotated data which are both costly and time consuming to obtain as human experts are often required to label those data. Moreover these datasets become quickly outdated due to the rapid evolution of new manipulation tools, aiming for continuous development of new highly annotated benchmarks, in a process which is both complex and expensive. Additionally, another non-trivial limitation lies in the notable amount of computational resources needed to train these models, which directly depends on their large number of trainable parameters. This makes their training heavily reliant on high-end hardware, thereby restricting their accessibility to users and stakeholders.
To address these limitations, we propose LADLE-MM (\emph{Limited Annotation based Detector with Learned Ensembles for Multimodal Misinformation}), a multimodal misinformation detection model specifically designed to operate under a limited annotation setting, i.e., without requiring explicit spatial or token-level grounding supervision. This design choice better reflects real-world deployment scenarios, where detailed annotations are costly, time consuming, and often unavailable, thereby enabling more scalable and practical misinformation detection systems.
Nonetheless, LADLE-MM achieves performance comparable to the state of the art while relying on limited data annotations and requiring fewer training resources. Our architecture fully exploits the benefits of contrastive learning and cross-modal attention, thanks to a three-branch design consisting of a text feature extraction, an image feature extraction, and a multimodal feature extraction branch. In the multimodal feature extraction branch we employ a single BLIP model as a fixed reference space, while transformer-based feature extractors are employed for the unimodal branches. The outputs from these branches are fused through a dual-path mechanism, with each path implemented as a cross-attention mechanism. This fusion strategy enables the model to effectively integrate information from different modalities and not overly rely just on a single modality. The resulting fused representations are subsequently employed to perform both binary and multi-label classification tasks; that is, to determine whether an image-text pair is pristine or tampered, and, in the latter case, to identify the specific types of manipulations that have been applied (see Section \ref{sec:experimental} for detailed definitions).

To improve generalization and performance, the architecture is pretrained on four distinct data splits of the DGM4 dataset \cite{dgm4}, following the model-soup protocol described by Liu et al. \cite{liu2024fka}. Subsequently, it is fine-tuned on the full DGM4 dataset using the model soup initialization strategy described by Wortsman et al. \cite{model_soup}, which averages the weights from the individual pretrained models, resulting in semi-ensembles of different models. 
This approach effectively approximates the performance benefits of ensembling without increasing inference costs, allowing us to obtain a lighter architecture in terms of trainable parameters, which enables training even on low-end hardware and also helps reducing the carbon footprint associated with training time. Despite this reduction, corresponding to a 60.3\% decrease in number of trainable parameters compared to the SOTA model \cite{dgm4}, we achieve comparable performance on both binary and multi-label classification tasks. Moreover, our model surpasses the best-performing model in the $mAP$ metric, highlighting its excellent ability to understand the various manipulations applied to the original image-text pairs. We also demonstrate the generalization abilities of our approach by evaluating it on the VERITE \cite{verite} dataset, where our model shows higher results than SOTA in an open-set setting. 

In summary, the main contributions of our work are the following:

\begin{itemize}
    \item We propose LADLE-MM, an efficient multimodal misinformation detection model which uses the model soup initialization strategy to improve robustness and boost performance. 
    
    \item We design a multimodal fusion mechanism that increases reliability and accuracy in misinformation detection, surpassing current state-of-the-art models in $mAP$ on the DGM4 benchmark when trained without grounding annotations, with 60.3\% fewer trainable parameters. 
    
    \item We enhance the fusion mechanism by adding a frozen multimodal branch, making attention correlate with manipulated regions, obtaining implicit image grounding capabilities and better generalization.
    
    \item  We perform extensive ablation tests to validate the necessity of a fixed multimodal reference space for robust cross-modal feature extraction. As part of this analysis, we compare multiple state-of-the-art multimodal backbones to justify our architectural design.
\end{itemize}

%% file: sec/2_related_work.tex
\section{Related Work}
Modern online content is inherently multimodal, making the robust fusion of heterogeneous signals a central challenge in representation learning. This problem has been extensively explored in fields like multimodal sentiment analysis and affective computing, where strategies such as prompt-learning \cite{fmsa, gpplea}, uncertainty-guided mechanisms \cite{feduaf, eupa}, and advanced interaction architectures \cite{raft, tempo, rmerdt, cime} have been developed to handle noisy or low-resource inputs. However, multimodal misinformation introduces uniquely adversarial constraints. Initially, research efforts focused mainly on detecting anomalies within a single modality, such as images or text alone. However, assessing veracity requires capturing subtle, often deliberate inconsistencies between modalities. Driven by the widespread diffusion of multimodal content on social media (such as image-text or video-text posts), the focus has shifted towards the detection of misinformation that involves generated data across multiple modalities. To contextualize this shift, the following subsections trace the evolution of misinformation detection from isolated unimodal approaches to unified multimodal strategies. 

\subsection{Misinformation Detection in Images}
With the emergence of powerful generative models such as GANs~\cite{gan-goodfellow2014}, images have become one of the most prominent media types used to convey misinformation. 
To address this issue, researchers initially focused on developing detectors for fake and manipulated images based on high-level features. 
These early works exploited specific image properties, such as color~\cite{color_components, detecting_gan}, saturation~\cite{detecting_gan}, lighting~\cite{lightning_inconsistencies}, and perspective~\cite{perspective_inconsistency}.
Other approaches rely on mesoscopic features~\cite{afchar2018mesonet} or global texture representations~\cite{liu2020global}. 
In this context, Ba et al.~\cite{exposing_deception} aimed to improve generalization by reducing overfitting to specific facial regions (such as eyes and mouth) through a disentanglement strategy applied to image patches. 
Similarly, Nataraj et al.~\cite{co_occurrency_detection} proposed using a color co-occurrence matrix as input for detecting fake and altered faces.
More recently, research has shifted towards low-level methods, extracting information from both the spatial~\cite{zhong2023patchcraft, tan2023learning, marra2019gans, yu2019attributing} and frequency domains~\cite{durall2020watch, 10156981, dzanic2020fourier, art12}, where generative models often leave detectable artifacts.
Finally, following the introduction of Vision Transformers (ViT)~\cite{liu2023survey}, attention-based mechanisms have been employed to detect altered images~\cite{dong2022protecting, zhuang2022uia, heo2021deepfake, wang2023deep}, thereby enhancing robustness and generalization.

\subsection{Misinformation Detection in Texts}
Detecting text-based fake news poses distinct challenges due to the intrinsic properties of language, which is highly variable and strongly influenced by the author’s style and intent. As a result, approaches to textual misinformation detection typically fall into two main categories: propagation-based and content-based methods. Propagation-based methods rely on features extracted from user profiles, the relationships among spreaders and their posts~\cite{profile_spreading, profile_spreading_2}, or the social media platform where the news is shared~\cite{social_network_spread}. 
However, a significant limitation of these approaches is that detection can only occur after the content has been published and shared. 
Consequently, since these methods rely on user interactions, they are unable to detect fake news before it has been widely shared. Conversely, content-based methods rely on the raw text itself to identify misinformation. 
Unlike propagation-based strategies, this approach enables the detection of fake news before it has been shared. 
However, this task is more challenging due to the previously mentioned complexities of natural language; specifically, it requires identifying the underlying and hidden intent within the text. Early approaches in this domain employed features extracted from text to perform classification using Support Vector Machines (SVM)~\cite{svm_fake_news}. 
Other methods focused on lexical analysis, utilizing techniques such as bag-of-words~\cite{bow_fake_news}, Part-of-Speech (POS) tagging~\cite{pos_fake_news}, and n-grams~\cite{ngram_fake_news}.

More recent strategies employ deep learning to analyze text and identify misinformation. 
Specific implementations include BiLSTM networks~\cite{bi-lstm-fake-news, bi-lstm-fake-news-2} and transformer-based architectures, such as BERT~\cite{bert-fake-news-1, bert-fake-news-2, bert-fake-news-3}. The latest advancements leverage Large Language Models (LLMs) enhanced with Retrieval-Augmented Generation (RAG) capabilities. 
These methods detect fake news in textual content by retrieving external information, thereby providing a broader context for verification~\cite{rag-fake-news-1, rag-fake-news-2}.

\subsection{Misinformation Detection in Multimodal Content}
The integration of image comprehension, generation, and editing capabilities in accessible large language models such as ChatGPT-4o \cite{chatgpt4o}, Gemini \cite{gemini}, and Grok 3 \cite{grok} has enabled a broad audience to generate highly realistic fake images from fabricated textual descriptions, or misleading textual descriptions from real images. This development significantly intensifies the threat posed by multimodal media to convey misinformation, especially on social media platforms and informational websites, where text and images frequently coexist to convey information \cite{art61, Atlam2025SLMDFS, LundbergEbba}. 

Initial efforts to detect multimodal misinformation relied on adversarial learning techniques, particularly focused on social network disinformation \cite{10.1145/3219819.3219903}, and bi-modal variational autoencoders \cite{10.1145/3308558.3313552}. More recently, the introduction of the Transformers \cite{vaswani} and, subsequently, Vision Transformers (ViT) \cite{art15} marked a turning point by enabling more sophisticated fusion mechanisms capable of jointly modeling image and text latent features. In SpotFake \cite{spotfake}, text features are extracted using a BERT model \cite{bert} and concatenated with image features extracted by a VGG-19 network \cite{vgg}, before being fed to a classifier. MSCA \cite{msca} adopts BERT and ViT as unimodal encoders, and employs a stacked co-attention mechanism to learn cross-modal interactions between text tokens and image patches.  Shao et al. \cite{dgm4} propose a hierarchical fusion strategy combined with a contrastive learning objective by developing their HAMMER architecture, which is later extended through cross-modality knowledge interaction in ViKI \cite{viki}, and gated attention mechanisms in VLP-GF \cite{vlp-gf}. The HAMMER architecture is further improved by enhancing the contrastive objective in HAMMER++ \cite{hammerpp}. To improve multimodal fusion, TT-BLIP~\cite{ttblip} introduces a tri-transformer architecture specifically designed to align visual and textual representations more effectively. Other approaches have explored the integration of frequency spectrum analysis for multimodal misinformation detection~\cite{frequency}, leveraging frequency-domain artifacts as complementary signals to visual and textual features.

Recent advancements have transitioned toward leveraging Large Vision-Language Models (LVLMs) for misinformation detection across diverse evaluation settings. This includes zero-shot prompt evaluation pipelines~\cite{jia2024chatgpt, tahmabi}, downstream instruction fine-tuning as demonstrated by SNIFFER~\cite{qi2024sniffer}, and targeted parameter fine-tuning strategies such as FKA-OWL~\cite{liu2024fka} and TRUST-VL~\cite{trustvl}. While these foundation frameworks inherently capitalize on extensive semantic world knowledge and generalize fluidly across unseen domains without architecture retraining, they introduce critical deployment bottlenecks. Specifically, their deployment real-world scenarios is constrained by massive computational latency, extreme sensitivity to prompt engineering, and a persistent risk of generating hallucinated outputs.
In contrast to these heavy architectures, LADLE-MM introduces a paradigm shift: rather than relying on prohibitively expensive and scarce dense annotations to supervise attention maps or visual bounding boxes, our approach encapsulates both global contrastive alignment and deep cross-modal attention within a parameter-efficient, three-branch layout anchored by a fully frozen reference space. In summary, while existing state-of-the-art methods rely heavily on extensive annotation supervision or massive foundation scale to capture inter-modal inconsistencies, LADLE-MM circumvents both resource constraints, achieving near-SOTA performance with a drastically optimized trainable parameter footprint.

%% file: sec/3_method.tex
\section{Method}
\begin{figure*}[t]
    \centering
    \includegraphics[width=\linewidth]{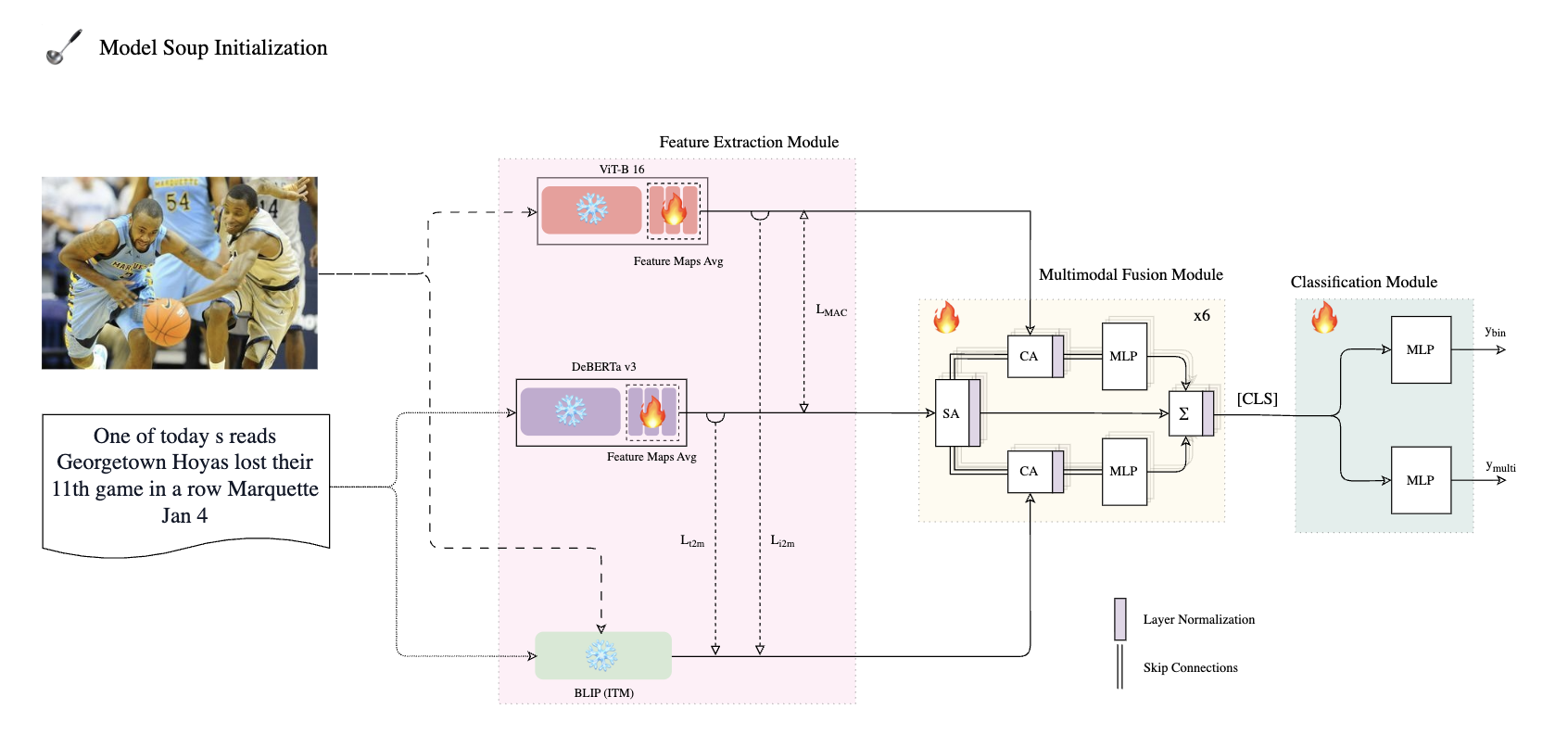}
    \caption{Overview of the proposed LADLE-MM: 1) Feature extraction module: \emph{ViT} for pure image feature extraction; \emph{DeBERTa} for pure text feature extraction; \emph{BLIP} for image-text pair feature extraction. 2) Multimodal feature fusion module consisting of six dual-path cross-attention blocks. 3) Binary and multi-label classification heads.}
    \label{fig:model-scheme}
\end{figure*}

To effectively tackle the multimodal misinformation detection in realistic low-annotation and resource-constrained settings, as shown in Figure \ref{fig:model-scheme}, we propose LADLE-MM: \emph{Limited Annotation based Detector with Learned Ensembles for Multimodal Misinformation}. LADLE-MM is a parameter-efficient architecture that leverages learned ensemble initialization and a fixed multimodal reference space to robustly identify cross-modal manipulations in image-text pairs. It can be summarized in three main modules:
\begin{itemize}
    \item \emph{Feature extraction module}: this module is composed of two \emph{uni-modal feature extraction branches} and a \emph{multimodal feature extraction branch}, aimed at creating a fixed reference space to facilitate the fusion of the two modalities.
    \item \emph{Multimodal feature fusion module}: this module is composed by six dual-path blocks where each path is composed of a multimodal cross-attention block, with textual modality as query and the image or multimodal embeddings as keys and values.
    \item \emph{Classification module}: this module consists of two separate classification heads performing binary and multi-label classification.
\end{itemize}

\subsection{Feature Extraction Module}
The uni-modal feature extraction branches enable the independent extraction of salient characteristics from images and texts, thereby allowing for a comprehensive understanding of intrinsic features for each modality independently. For each encoder, we aggregate information by averaging the outputs of the final layers, which balances semantic richness and robustness while limiting additional trainable parameters.

The image feature extraction branch is based on a pre-trained \emph{Vision Transformer (ViT)} \cite{art15}, where only the last three layers of the encoder are fine-tuned. The output is a feature tensor that includes a classification token and a latent representation for each non-overlapping patch of the input image. This design allows the model to capture both a global representation of the image through the classification token and local information from individual patches, making it possible to detect manipulations at both the general and detailed levels.

The text feature extraction branch relies on a pre-trained language model, with only the last three layers of the model fine-tuned during training. It enables the extraction of latent features from the text modality alone, providing both a global representation of the input and detailed representations for each individual token. The output of this branch is used as the input signal to be conditioned in the fusion module, which is later employed for the final classification.

The multimodal feature extraction branch is based on a pre-trained, fully frozen Vision Language Model (VLM), equipped with an image-text matching head. Image-text pairs are jointly fed into the model to extract a unified representation that captures cross-modal interactions between modalities. This not only enriches the information obtained from each modality individually, but also provides a stable, shared latent space that facilitates a more effective fusion of the two modalities.

Features extracted from the two uni-modal branches are then aligned using the \emph{Manipulation-Aware Contrastive Loss (MAC Loss)}~\cite{dgm4}, with a reduced queue of negative examples. 

The negative sample queue represents a critical component of the implemented loss function. 
Functionally, it acts as a dynamic dictionary of previously processed samples specifically, mismatched pairs or misinformation content, effectively extending the standard InfoNCE formulation. By decoupling the number of negative samples from the mini-batch size, the queue allows for a massive expansion of the contrastive pool, enabling the model to learn more discriminative representations and achieve better and more discriminative alignment in the embedding space. 

However, the use of a queue can introduce training instability due to feature inconsistency: since the encoder parameters evolve during training, the stored representations (keys) may become outdated compared to the current queries. 
To mitigate this distribution shift, we employ a \emph{Momentum Encoder} i.e. a copy of the base encoder whose weights are updated via an Exponential Moving Average (EMA). 
This mechanism ensures that the key representations in the queue evolve smoothly, maintaining consistency across training iterations and preventing rapid fluctuations in the latent space.

More Formally, let $z_\text{img, CLS}$ and $z_\text{txt, CLS}$ denote the classification tokens of the image and text representations in a selected batch, respectively. Let $z_\text{img}^+$ and $z_\text{txt}^+$ be the corresponding original (i.e., non-manipulated) image and text pairs representations extracted with a momentum encoder, and $I_Q$ and $T_Q$ the queues of negative samples, composed of past and manipulated data representations. The individual loss components are defined as:
\begin{align}
    & \mathcal{L}_{i2t} = \text{InfoNCE}(z_\text{img, CLS}, z_\text{txt}^+, T_Q) \\
    & \mathcal{L}_{t2i} = \text{InfoNCE}(z_\text{txt, CLS}, z_\text{img}^+, I_Q) \\
    & \mathcal{L}_{i2i} = \text{InfoNCE}(z_\text{img, CLS}, z_\text{img}^+, I_Q) \\
    & \mathcal{L}_{t2t} = \text{InfoNCE}(z_\text{txt, CLS}, z_\text{txt}^+, T_Q)
\end{align}
where InfoNCE \cite{Oord2018RepresentationLW} is a type of contrastive learning loss that encourages the model to assign high similarity scores to positive pairs and low similarity scores to mismatched pairs within the batch, defining the task as a classification objective where the positive pair represents the correct class. In this work, it has been adapted to handle longer sequences of negative pairs by using a queue of negatives handled by a momentum encoder as in He et al. \cite{moco_loss}, allowing for better discriminative abilities by the model. A definition of the implemented loss is the following:
\begin{equation}
    \text{InfoNCE}(z, z^+, Q) = -\log\underset{z'\in\{z^+\} \cup Q}{\text{softmax}}\left[\frac{d(z, z')}{\tau}\right]_{z'=z^+}
    \label{eq:infonce}
\end{equation}
where $z$ is the query, $z^+$ is the positive match, $Q$ is a queue of negative matches, $\tau$ the temperature, and $d(\cdot, \cdot)$ cosine similarity.
The final MAC Loss is computed as the average of the four components:
\begin{equation}
    \mathcal{L}_\text{MAC} = \frac{1}{4} \left(
    \mathcal{L}_{i2t} + \mathcal{L}_{t2i} + \mathcal{L}_{i2i} + \mathcal{L}_{t2t}
    \right)
\end{equation}
This loss encourages the model to bring closer the representations of aligned, non-manipulated image-text pairs, while pushing apart those in which one or both modalities are manipulated.
In addition to this, uni-modal features are aligned with the stable representations extracted by the multimodal encoder. Let $z_\text{multi, ENC}$ be the \texttt{ENCODE} token from the multimodal representation. The cosine similarity alignment losses are defined as:
\begin{align}
    & \mathcal{L}_{i2m} = 1 - \frac{z_\text{img, CLS} \cdot z_\text{multi, ENC}}{\|z_\text{img, CLS}\| \|z_\text{multi, ENC}\|} \\ 
    & \mathcal{L}_{t2m} = 1 - \frac{z_\text{txt, CLS} \cdot z_\text{multi, ENC}}{\|z_\text{txt, CLS}\| \|z_\text{multi, ENC}\|} 
\end{align}
and their combined form:
\begin{equation}
    \mathcal{L}_\text{cos-similarity} = \frac{1}{2} \left( \mathcal{L}_{i2m} + \mathcal{L}_{t2m} \right)
\end{equation}

The final alignment objective combines the contrastive and cosine similarity terms as:
\begin{equation}
    \mathcal{L}_\text{align} = \mathcal{L}_\text{MAC} + \lambda_c \mathcal{L}_\text{cos-similarity}
\end{equation}
where $\lambda_c$ denotes a scalar hyperparameter that balances the contribution of the loss term, empirically set to $0.3$ in our implementation. 

This combined objective extends the method proposed by Li et al. \cite{dgm4}, integrating a cosine similarity loss $\mathcal{L}_\text{cos-similarity}$ to produce smoother fusion of semantically similar representations through the fixed reference space provided by the multimodal feature extraction branch. 

The intuition behind integrating the $\mathcal{L}_\text{cos-similarity}$ alongside the contrastive objective is that a shared space ensures semantic coherence across modalities, grounding the learnable uni-modal features within a semantically stable manifold.  By minimizing the distance between the uni-modal classification tokens ($z_\text{img, CLS}$, $z_\text{txt, CLS}$) and the frozen multimodal reference $z_\text{multi, ENC}$, we impose a "semantic anchor" on the training process. This fixed reference space ensures that the image and text encoders do not drift into disjoint feature spaces optimized solely for the contrastive task, but instead converge into a shared, cohesive region defined by the VLM. Furthermore, by aligning the representations within the same semantic region, the model reduces its reliance on a single modality, ensuring a balanced fusion.

\subsection{Multimodal Feature Fusion Module}
Each block of the proposed \emph{multimodal fusion module} consists of two multimodal cross-attention branches, both enhanced with skip connections. At the beginning of each block, the text features inputs are first processed through a Self-Attention (SA) mechanism. The first branch performs fusion between the text and image modalities, while the second branch fuses text with the multimodal features. Both branches include a Multi-Head Cross-Attention (CA) layer followed by a Multi-Layer Perceptron (MLP), each equipped with Layer Normalization (LN) and residual (skip) connections. For both branches, text features represent the query for the attention mechanism.

The outputs of the two fusion branches are then added together with the residual self-attended text representation, producing the final fused output of the block.

Let $z_\text{txt}$, $z_\text{img}$, and $z_\text{multi}$ be the latent features for text, image, and text-image respectively.
The self attended text representation can be represented as:
\begin{equation}
    z_t = \text{LN}(\text{SA}(z_\text{txt}) + z_\text{txt})
\end{equation}
The first branch can be represented as:
\begin{align}
& z_\text{it} = \text{LN}(\text{CA}_\text{txt-img}(z_t, z_\text{img}) + z_t) \\
& z_\text{txt-img} = \text{LN}(\text{MLP}_\text{txt-img}(z_\text{it}) + z_\text{it})
\end{align}
The second branch can be represented as:
\begin{align}
& z_\text{tm} = \text{LN}(\text{CA}_\text{txt-multi}(z_t, z_\text{multi}) + z_t) \\
& z_\text{txt-multi} = \text{LN}(\text{MLP}_\text{txt-multi}(z_\text{tm}) + z_\text{tm})
\end{align}
A single block output can be represented as:
\begin{equation}
    z_\text{fused} = \text{LN}(z_\text{txt} + z_\text{txt-multi} + z_\text{txt-img})
    \label{eq:fused}
\end{equation}

The complete fusion module is composed of six consecutive blocks of this structure. We experimentally determined this to be the optimal depth, as demonstrated in the ablation study and Table \ref{tab:ablation}.
This design integrates multimodal joint embeddings with unimodal embeddings derived from both images and text and produces a fused, semantically meaningful representation of the input image-text pair.

\subsection{Classification Module}
The classification module consists of two separate heads: one for binary classification and one for multi-label classification.
Both heads take as input the fused classification token $z_\text{fused}$, as defined in Eq.~\ref{eq:fused}. Each head is implemented as a simple feed-forward network. The binary classification head outputs a single score indicating the likelihood of falseness, while the multi-label classification head produces four independent outputs, corresponding to the presence of the following possible manipulations: Face Swap (FS), Face Attribute (FA) manipulation, Text Swap (TS), and Text Attribute (TA) manipulation.
Both the binary classification task and the multi-label classification task are optimized using a Binary Cross-Entropy (BCE) loss with logits. Let $\mathcal{L}_{bin}$ denote the binary classification loss and $\mathcal{L}_{multi}$ denote the multi-label classification loss. 
The final loss $\mathcal{L}_{total}$ used to train LADLE-MM is defined as the weighted sum of these two components and the alignment objective $\mathcal{L}_{align}$ (defined in Section 3.1):

\begin{equation}
    \mathcal{L}_{total} = \alpha \mathcal{L}_{align} + \mathcal{L}_{bin} + \mathcal{L}_{multi}
\end{equation}

where $\alpha$ is a scaling parameter used to balance the contribution of the alignment loss with the classification losses. We empirically identifies $\alpha = 0.1$ to be the optimal value, prioritizing the classification task while still leveraging the semantically cohesive latent space established by the alignment losses.

%% file: sec/4_experiments.tex
\section{Experimental Setup}
\label{sec:experimental}
\begin{figure}
    \centering
    \includegraphics[width=\linewidth]{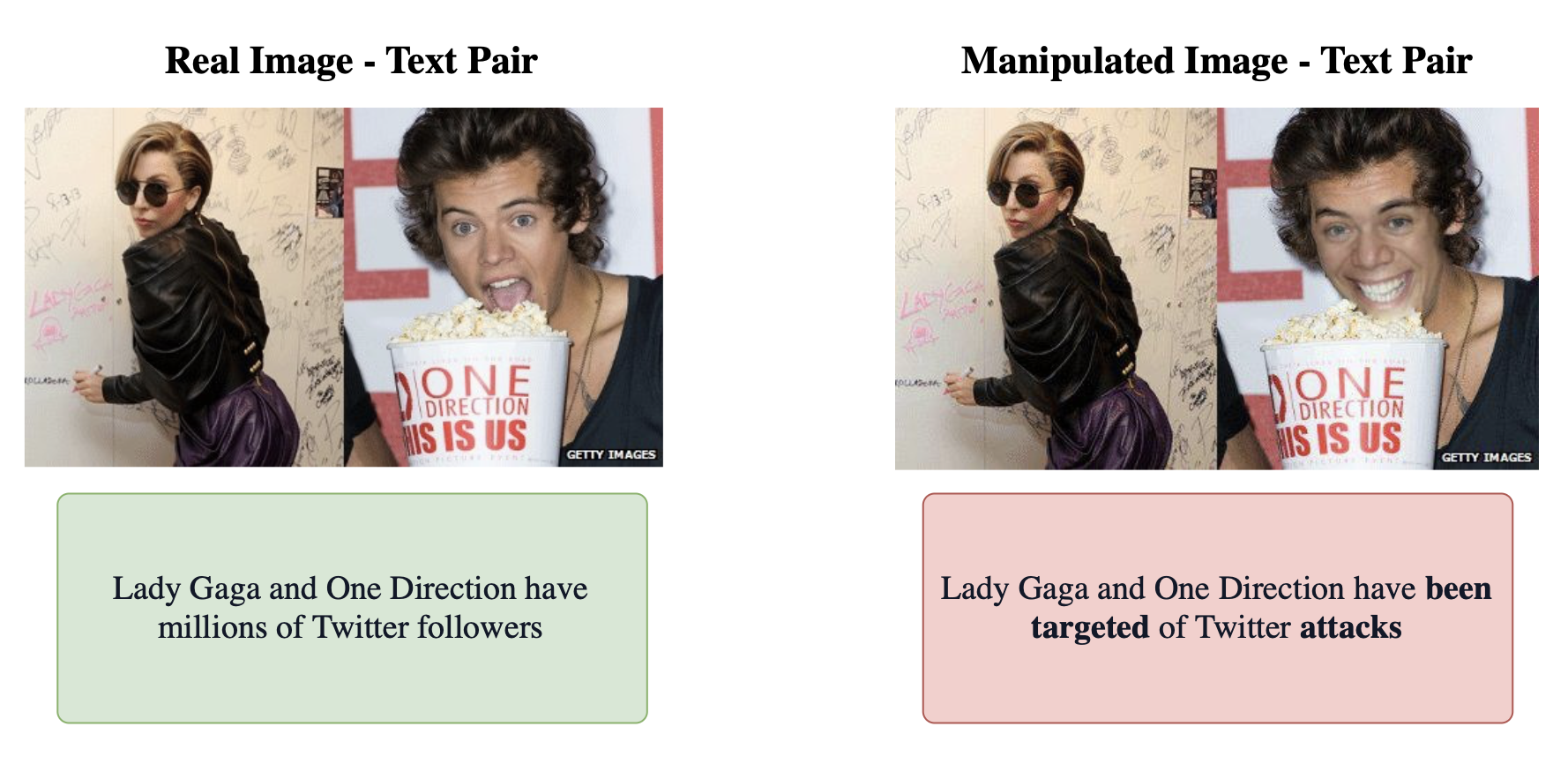}
    \caption{On the left, an original image-text pair retrieved from the VisualNews dataset; on the right, the corresponding manipulated sample extracted from DGM4.}
    \label{fig:unmanipulated_pair}
\end{figure}

\paragraph{Datasets}
Our model has been trained and evaluated on the DGM4 benchmark~\cite{dgm4}. This dataset is designed specifically for multimodal misinformation detection involving both image and text forgeries. It comprises real and fake image-text pairs, each drawn from four different news outlets: BBC, The Guardian, USA Today and The Washington Post. Those different splits manifest specific linguistic features and manipulation patterns characteristic of each source, reflecting differences between British and American English as well as in the tampering techniques applied. Each split contains data exclusively from one outlet, as released by Liu et al.~\cite{liu2024fka}\footnote{\url{https://github.com/liuxuannan/FAK-Owl}}. DGM4 also provides annotations for binary classification, multi-label classification, manipulated text token grounding, and bounding boxes for manipulated regions. In our work, we focus only on binary classification (i.e., distinguishing real from fake image-text pairs) and multi-label classification, with the following categories:
\begin{itemize}
    \item Face Attribute Manipulation (FA) involves modifying facial features, expressions, or emotions without changing the subject identity.
    \item Face Swap (FS) entails replacing a face within the image, effectively altering the identity of the depicted person.
    \item Text Attribute Manipulation (TA) consists of substituting specific words to shift the sentiment or emotional tone of the text.
    \item Text Swap (TS) involves substituting or recombining words to alter the semantic meaning, while preserving the primary character or entity mentioned in the text.
\end{itemize}
Additionally, the \emph{VisualNews} dataset~\cite{visual_news} has been used as a support dataset to retrieve the original (i.e., unmanipulated) image-text pairs, which are necessary to compute the manipulation-aware contrastive loss. An example of a manipulated and its corresponding original image-text pair is shown in Fig.~\ref{fig:unmanipulated_pair}.
To evaluate the generalization capabilities of the proposed method we employ VERITE~\cite{verite}, a real-world benchmark created to ensure a robust and reliable evaluation framework and effectively address unimodal bias, rendering it a solid evaluation framework for Multimodal Misinformation Detection.
This design ensures a robust evaluation framework, allowing to assess both the model adaptability to unseen domains and its ability to detect misinformation without disproportionate reliance on a single modality.
Assessing resistance to unimodal bias is critical for misinformation detection systems. 
If a model relies disproportionately on a single modality, even minor shifts or perturbations in that modality can lead to significant performance degradation. 
Furthermore, while such over-reliance might yield high performance on in-domain data, it often results in catastrophic failure during out-of-distribution evaluations. 
Moreover, in the specific context of misinformation detection, focusing solely on one modality increases the risk of false positives or negatives. 
These errors could otherwise be prevented by effectively leveraging the supporting evidence provided by the complementary modality.

\begin{table}[!ht]
    \centering
    \begin{tabular}{l|c}
    \toprule
    \textbf{Methods}
     & \textbf{\# Trainable Parameters ($\downarrow$)} \\
    \midrule
    HAMMER \cite{dgm4} & 228 M \\
    $\text{Viki}^\dagger$ \cite{viki} & $\sim$ 228 M \\
    $\text{VLP-GF}^\dagger$ \cite{vlp-gf} &  $\sim$ 228 M \\ 
    $\text{TT-BLIP}^\dagger$ \cite{ttblip} & $\sim$ 250 M \\
    \textbf{LADLE-MM (Ours)} & \textbf{89 M}
    \\
    \bottomrule
    \end{tabular}
    \caption{Number of trainable parameters for the models evaluated on DGM4. Approaches marked with $\dagger$ do not provide an official release, so the reported numbers are approximate. For Viki and VLP-GF, we assume the same number of parameters as HAMMER, since they are both based on it. For TT-BLIP, we estimate a parameter count similar to ours in the feature extraction branch, but higher overall due to the use of three BLIPs. Our approach has 60.3\% fewer trainable parameters compared to HAMMER.}
    \label{tab:parameters}
\end{table}

\begin{table*}[!h]
    \centering
    \begin{tabular}{l|ccc|ccc|c|c}
    \toprule
    \textbf{Methods} & \multicolumn{3}{c|}{\textbf{Binary Cls}} & \multicolumn{3}{c|}{\textbf{Multi-Label Cls}} & \textbf{Reduced Annotations} & \textbf{Model Soup} \\
     & AUC $\uparrow$ & ACC  $\uparrow$ & EER $\downarrow$ & mAP $\uparrow$ & CF1 $\uparrow$ & OF1 $\uparrow$ \\
    \midrule
    CLIP~\cite{art21} & 83.22 & 76.40 & 24.61 & 66.00 & 59.52 & 62.31 & \XSolidBrush & \XSolidBrush \\
    ViLT~\cite{art22} & 85.16 & 78.38 & 22.81 & 72.37 & 66.14 & 66.00 & \XSolidBrush & \XSolidBrush \\
    VLP-GF$^\dagger$ \cite{vlp-gf} & 92.84 & 86.13 & 14.45 & 85.65 & 80.02 & 79.07 & \XSolidBrush & \XSolidBrush\\
    HAMMER++$^\dagger$ \cite{hammerpp} & 93.33 & 86.39 & 14.10 & 86.41 & 79.73 & \cellcolor{blue!25}80.71 & \XSolidBrush & \XSolidBrush \\
    ViKI$^\dagger$ \cite{viki} & \cellcolor{blue!25}93.51 & \cellcolor{blue!25}86.67 & \cellcolor{blue!25}13.87 & 86.58 & \cellcolor{blue!25}81.07 & 80.10 & \XSolidBrush & \XSolidBrush\\
    \midrule
    HAMMER$^\ddagger$ \cite{dgm4} & 88.65 & 80.26 & 19.83 & 79.85 & 70.41 & 70.53 & \Checkmark & \XSolidBrush\\
    \textbf{LADLE-MM (Ours)} & \textbf{92.20} & \textbf{83.84} & \textbf{16.23} & \cellcolor{blue!25}\textbf{87.90} & \textbf{78.46} & \textbf{76.84} & \Checkmark & \Checkmark \\
    \bottomrule
    \end{tabular}
    \caption{Comparison between LADLE-MM and other approaches on the DGM4 dataset. Performance is evaluated using Accuracy (ACC), Area Under the ROC Curve (AUC), and Equal Error Rate (ERR) for binary classification, as well as mean Average Prevision (mAP), average per-class F1 (CF1), and average overall F1 (OF1) for fine grained multi-label manipulation detection. Symbols $\uparrow$ and $\downarrow$ denote whether higher or lower values indicate better performance, respectively. Valued shaded in \colorbox{blue!25}{lilac} represent the absolute best performace across all methods. Values in \textbf{bold} highlight the best performance achieved under the reduced annotations regime (i.e. model trained without explicit spatial token-level grounding). Models marked with $^\dagger$ do not publicly release their code, while the models marked with $\ddagger$ does, enabling evaluation with limited annotations. In the column denoted by "Reduced Annotations" it is specified whether the models are trained without the use of grounding labels, while the column "Model Soup" specifies whether the model soup approach has been used during training.}
    \label{tab:results}
\end{table*}

\subsection{Architecture and Model Soup Training Protocol} 
The first stage of the architecture is the \emph{Feature Extraction Module}, comprising of two unimodal branches and a multimodal branch.
For the image feature extraction branch we employ a pretrained ViT-base model with a patch size of 16 and input resolution of 224 as image features extractor, specifically we use \texttt{google/vit-base-patch16-224} \footnote{\url{https://huggingface.co/google/vit-base-patch16-224}} from HuggingFace. For text feature extraction, we employ the pretrained DeBERTa-v3 \cite{art16} base model, namely \texttt{microsoft/deberta-v3-base} \footnote{\url{https://huggingface.co/microsoft/deberta-v3-base}}. In both cases, only the last three layers of each encoder are updated during the pretraining phase, while the entire encoder is kept frozen during fine-tuning. Lastly we adopt BLIP \cite{art17} for image-text matching as multimodal feature extractor, since this model has consistently demonstrated strong alignment capabilities between visual and textual content~\cite{blip_1, blip_3}. Our preliminary experiments and subsequent ablation studies (see Section \ref{sec:ablation}) confirmed that BLIP provides a more stable multimodal reference space compared to other configurations. Specifically we use the \texttt{Salesforce/blip-itm-base-coco} \footnote{\url{https://huggingface.co/Salesforce/blip-itm-base-coco}} model pretrained on the COCO dataset. This module is kept fully frozen throughout both pretraining and fine-tuning. All three encoders output 768-dimensional hidden representation, allowing the representation to operate within a common embedding space. This shared embedding dimensionality enables direct feature interaction within the proposed cross-attention fusion module without introducing additional projection layers, contributing to the parameter-efficient design of LADLE-MM.

The second stage of the architecture features the \emph{Multimodal Fusion Module}, which is composed of a sequentially stacked sequence of six identical fusion blocks. Within each block features are first passed through self-attention layer. The output then splits into a a dual-path cross-attention mechanism utilizing features extracted from both the images and the multimodal layer. Finally, the parallel representations are passed through a multi-layer perceptron (MLP) block and summed. While all six fusion blocks are fully trainable during the pretraining stage, only the last three fusion blocks remain trainable during the model soup fine-tuning. The fused representations are then passed to the \emph{Classification Module} that extracts the [CLS] token to perform multi-task inference with two distinct heads.
In order to obtain a robust and generalized multimodal misinformation detector, we directly exploit the diversified nature of the DGM4 benchmark by pretraining LADLE-MM independently on the four outlet-specific splits. After this pretraining phase, the weights from each of the four split-specific models are extracted and subsequently averaged. This averaged model is then employed as the initialization for a final fine-tuning stage on the full DGM4 dataset, following the \emph{model soups} strategy proposed by Wortsman et al.~\cite{model_soup}. 
We leverage this multi-stage initialization strategy to directly target the intrinsic domain diversity characteristic of multimodal misinformation detection. By pretraining four different instances of LADLE-MM, we create a set of models that have perfectly learned the nuances of the different outlet-specific splits. 
The final merging and fine-tuning phase allows LADLE-MM to internalize the outlet-specific knowledge into a single model with improved robustness and generalizability, enhancing the model ability to detect sophisticated manipulations across modalities without increasing either the inference time or the architecture dimensionality.
To the best of our knowledge, this is the first time that such an approach has been applied to the task of multimodal misinformation detection. During the pretraining stage, the $\mathcal{L}_{MAC}$ contrastive loss is employed 
to enhance the quality of the extracted features by emphasizing the distinction between positive and negative pairs. This loss function utilizes a negative queue of $2048$ samples, a temperature parameter of $\tau=0.05$, and a momentum of $\mu=0.999$.
Overall, our model includes only 89M trainable parameters during pretraining and just 33.9M during fine-tuning. This makes it significantly lighter to train compared to HAMMER (228M trainable parameters) and its derivatives~\cite{vlp-gf, viki, hammerpp, art24}, corresponding to a 60.3\% reduction in trainable parameters. Furthermore, our approach requires fewer parameters than TT-BLIP~\cite{ttblip}, as we do not rely on a dedicated BLIP model for each branch. A detailed comparison of trainable parameter counts is provided in Table~\ref{tab:parameters}.

\begin{figure}[!h]
    \centering
    \includegraphics[width=\linewidth]{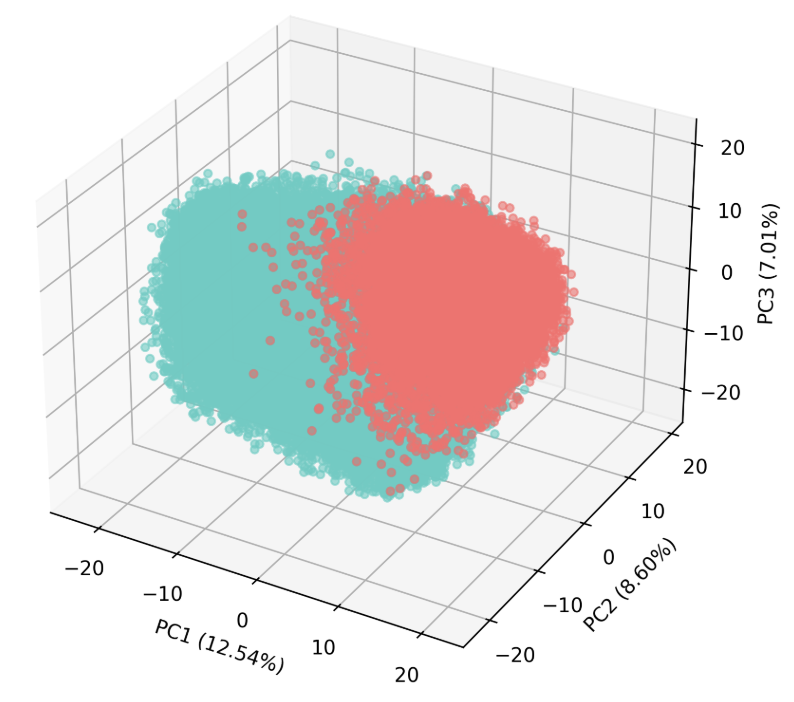}
    \caption{A three-dimensional Principal Component Analysis (PCA) visualization of the classification tokens extracted from the fusion module. The axes represent the three most significant directions of variance in the data: PC1 (capturing 12.54\% of the variance), PC2 (8.60\%), and PC3 (7.01\%).}
    \label{fig:pca}
\end{figure}

We train LADLE-MM on 4 RTX ADA 5000 GPUs for 50 epochs, adopting early stopping to prevent overfitting. The process employs an effective batch size of $512$, obtained via a per-GPU batch size of $32$ and $16$ gradient accumulation steps. Optimization is handled by AdamW with a learning rate of $10^{-4}$ and a selective weight decay of $0.02$ applied only to linear and convolutional layers. We utilize a Cosine Annealing learning rate schedule ($T_{max}=5$, $\eta_{min}=10^{-7}$). Following the initialization of the final weights via the Model-Soup protocol, the architecture undergoes a general fine-tuning phase for 20 epochs. Both stages are executed using PyTorch Lightning with \texttt{bf16-mixed} precision and gradient clipping set to $1.0$.

Analytical evaluation of the computational complexity shows that a single forward pass requires approximately 148.44 GFLOPs per sample. The majority of this computation is handled by the frozen encoders (ViT-Base: 35.13 GFLOPs, DeBERTa-Base: 29.90 GFLOPs, and BLIP: 67.60 GFLOPs). By freezing these base models during fine-tuning, the backward pass is restricted solely to the trainable modules (the 6-block Fusion Layer and the Classifiers), requiring only 31.62 GFLOPs. Consequently, the total training cost is reduced to roughly 180.06 GFLOPs per sample, a substantial decrease compared to the estimated 445 GFLOPs needed if the entire architecture were unfrozen.
This theoretical efficiency is directly supported by empirical measurements. Tracking indicates an average training time of 5933.22 seconds per epoch for our architecture. In contrast, the Hammer baseline requires 7788.20 seconds per epoch, demonstrating the clear computational advantage of our partially frozen design.

%% file: sec/5_results.tex
\section{Results}
In this section we present a comprehensive evaluation of LADLE-MM across different settings to evaluate its robustness and efficiency. First, we analyze its performance on the DGM4 benchmak for both binary and multi-label classification tasks, comparing it against existing state-of-the-art approaches. Later we further evaluate it on the VERITE dataset, in order to assess the model ability to perform in an out-of-distribution scenario. Finally, we report the results of extensive ablation studies to quantify the contributions of key architectural components and design choices.

\subsection{Binary and Multi-Label Classification}
Following related works in the field of multimodal misinformation detection, as evaluation metrics we adopt Accuracy (ACC), Area Under the Receiver Operating Characteristic Curve (AUC), and Equal Error Rate (EER) for binary classification, and  mean Average Precision
(mAP), average per-class F1 (CF1), and average overall F1 (OF1) for evaluating the detection of fine-grained manipulation types.
To ensure a rigorous evaluation, Table \ref{tab:results} presents comparisons under two different scenarios: Full Supervision (where models leverage dense, auxiliary text-token and image-bounding-box grounding annotations) and Reduced Annotations (where models operate solely on image-text pairs without localization labels).
Under identical training conditions within the reduced annotation regime, LADLE-MM substantially outperforms the baseline HAMMER framework across all metrics.
Remarkably, even when compared against highly complex state-of-the-art architectures that benefit from full grounding supervision, our parameter-efficient model achieves highly competitive performance, trailing by only 1.31 points in AUC while securing a 1.32-point improvement in multi-label mAP.
To better understand how the model makes its predictions we analyzed how it positions in its latent space the representations of different multimodal inputs, performing a three-dimensional PCA \cite{art23} on the final classification tokens. As shown in Fig.~\ref{fig:pca}, the model correctly separates in the latent space the embeddings of real and fake content, demonstrating a good understanding of the underlying differences between the features that discriminate authentic and synthetic text-image pairs. 

\begin{figure}[htb]
    \centering
    \includegraphics[width=\linewidth]{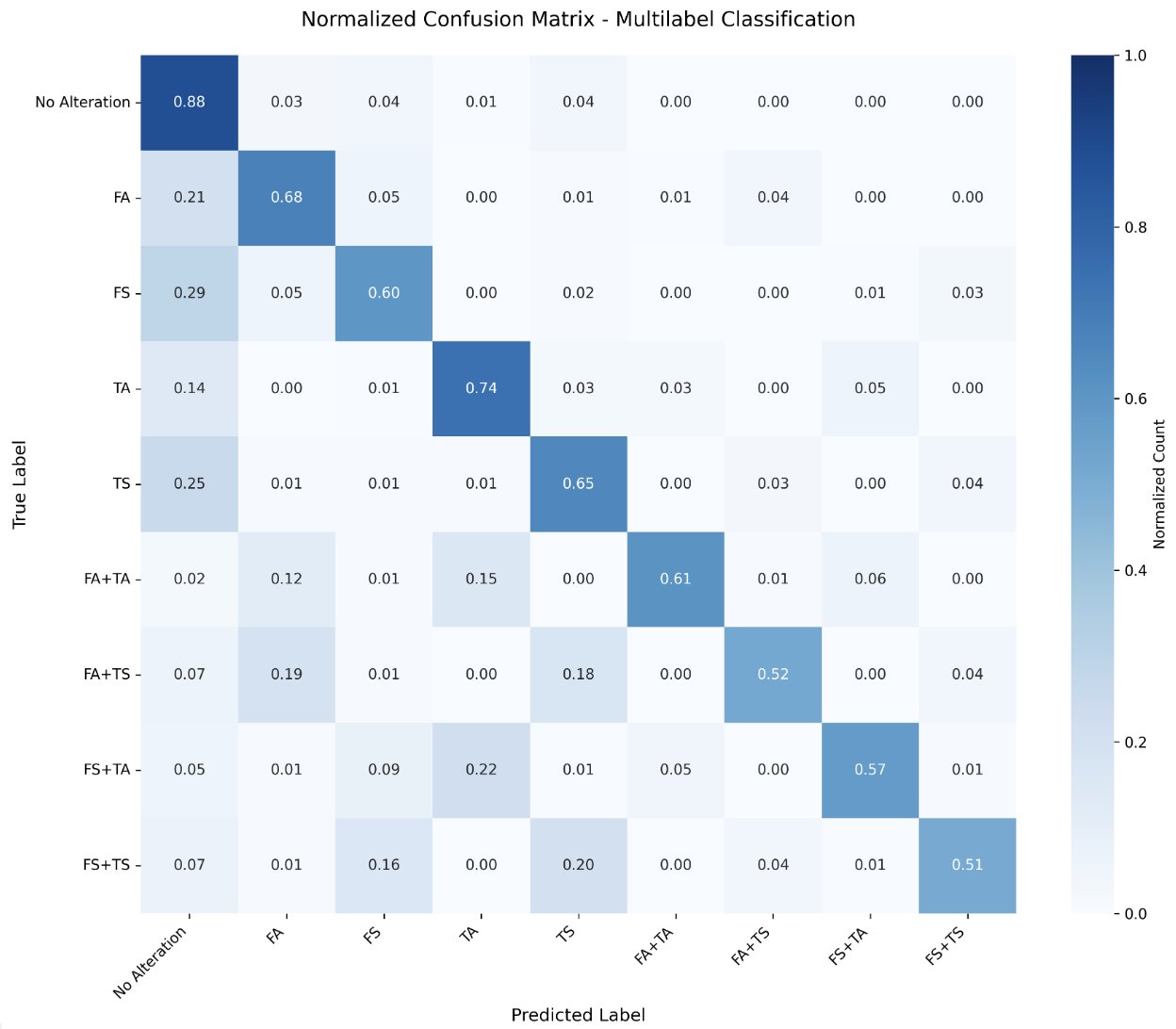}
    \caption{Confusion matrix for the multi-label classification task. 
The visualization highlights that misclassifications primarily occur between manipulation classes and the ``no alteration'' category, rather than confusing distinct manipulation types with each other, demonstrating the model discriminative robustness.}
    \label{fig:confusion_matrix}
\end{figure}

\begin{figure}[!ht]
    \centering
    \includegraphics[width=\linewidth]{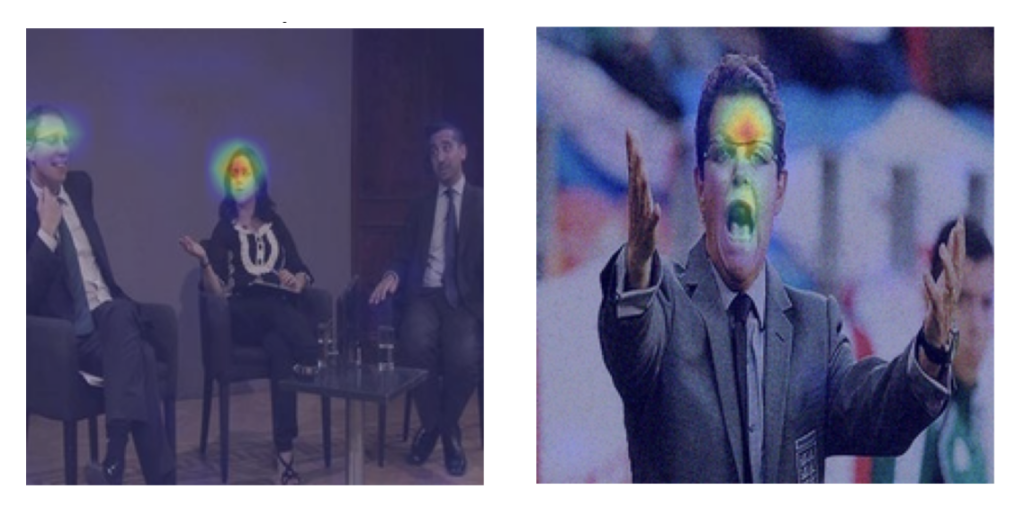}
    \caption{Attention map obtained using the \emph{beyond attention} method on the cross-attention layers of the text-image branch. The highlighted areas show where the model focuses to detect manipulations.}
    \label{fig:attention_map}
\end{figure}

To further examine the behavior of the proposed LADLE-MM in the multi-label classification setting, we report the model confusion matrix in Fig.~\ref{fig:confusion_matrix}.
As illustrated, errors in single manipulations predominantly correspond to prediction of the pristine class, rather than confusions among different manipulation types. Moreover, in instances containing combined manipulations, even when the model fails to predict the exact label combination, it typically correctly identifies at least one of the two manipulations. 
This behavior indicates that the model is not only robust but also capable of reliably distinguishing between diverse forms of image-text alterations.
\begin{figure*}[htbp]
    \centering
    \includegraphics[width=\linewidth]{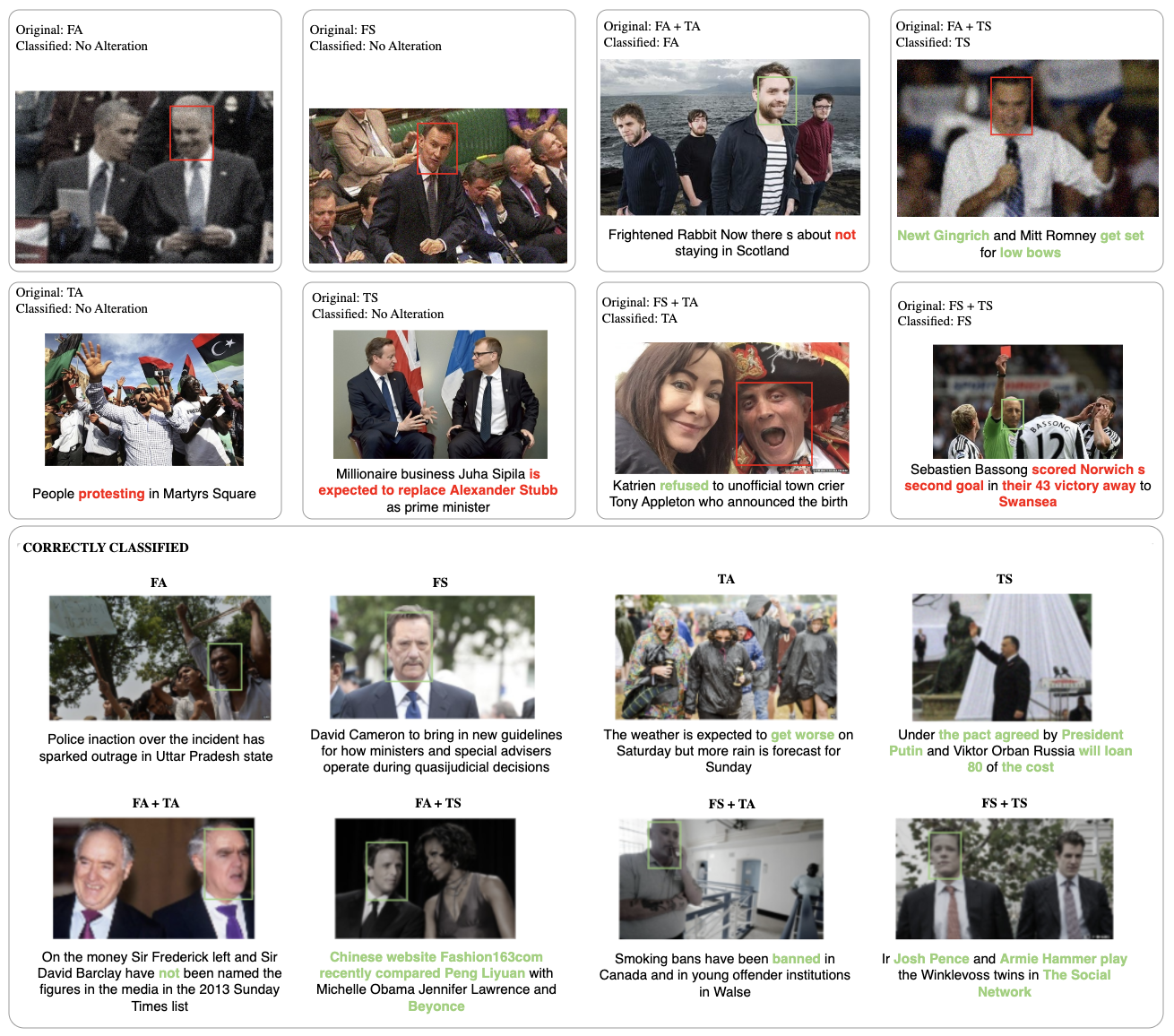}
\caption{Qualitative examples of misclassified and correctly classified samples. 
Visual misclassifications often arise from low-quality inputs or cluttered scenes, which are inherently ambiguous even for humans. (\textit{FA = Face Attribute manipulation, FS = Face Swap, TA = Text Attribute manipulation, and TS = Text Swap.}) Conversely, textual errors primarily stem from the lack of external knowledge required to verify specific named entities or events. Correct predictions are highlighted in green, while mistakes in red. The bounding boxes are provided solely for visual reference and are not model predictions, as LADLE-MM produces only class labels without spatial grounding.}
    \label{fig:misclassification}
\end{figure*}

Qualitative examples of misclassifications are illustrated in Fig.~\ref{fig:misclassification}. It can be clearly observed that visual errors predominantly occur in low-quality images or in scenes containing multiple subjects (crowded scenes). 
Conversely, regarding the textual modality, the majority of errors stem from the model inability to access external resources, which prevents the verification of specific named entities or events.
\begin{table}[!h]
    \centering
    \resizebox{\columnwidth}{!}{%
    \begin{tabular}{lcc|c}
    \toprule
    \textbf{Methods} & \multicolumn{2}{c}{\textbf{VERITE}} & \textbf{EK} \\
     & ACC & F1 & \\
    \midrule 
    BLIP2 \cite{li2023blip} & 50.75 & 37.35 & \XSolidBrush \\
    InstructBLIP \cite{dai2023instructblip} & 52.50 & 49.60 & \XSolidBrush \\
    LLaVA \cite{liu2023visual} & 52.75 & 49.80 & \XSolidBrush \\
    xGen-MM \cite{xue2024xgen} & 54.50 & 54.41 & \XSolidBrush \\
    LLaVA-NeXT \cite{liu2024llavanext} & 54.75 & 54.57 & \XSolidBrush \\
    Qwen2-VL \cite{wang2024qwen2} & 70.00 & 68.94 & \XSolidBrush\\
    GPT-4o \cite{chatgpt4o} & 67.50 & 67.57 & \Checkmark \\
    o1 \cite{openai2024o1} & 71.75 & 71.66 & \Checkmark \\
    \midrule
    MMD-Agent \cite{liu2024mmfakebench} & 54.75 & 47.00 & \Checkmark \\
    SNIFFER \cite{qi2024sniffer} & 72.50 & 72.02 & \Checkmark \\
    LRQ-FACT \cite{beigi2025can} & 64.75 & 68.32 & \Checkmark \\
    TRUST-VL \cite{yan2025trust} & 73.75 & 73.61 & \Checkmark \\
    \midrule
    \textbf{LADLE-MM (Ours)} & \textbf{79.60} & \textbf{83.50} & \XSolidBrush \\
    \bottomrule
    \end{tabular}%
    }
    \caption{Out-of-distribution results on the VERITE dataset, compared to general purpose VLM and misinformation detectors. LADLE-MM obtains the best results, showing high resistance to unimodal bias and outperforming large-scale models. The column denoted with ``EK'' specifies whether the approaches uses an external knowledge to make its predictions.}
    \label{tab:ood_results}
\end{table}
Even without relying on a dedicated grounding module to explicitly localize manipulated patches, our method effectively learns to focus on relevant regions to detect image manipulations. In particular, the text-image cross-attention blocks concentrate on areas like the eyes and mouth, which are common sites for manipulations. This behavior can be visualized using the \emph{beyond attention} method proposed by Chefer et al.~\cite{beyond_attention}, as shown in Fig.~\ref{fig:attention_map}.

\begin{table*}[!ht]
    \centering
    \begin{tabular}{l|cc|ccc}
    \toprule
    \textbf{Multimodal Backbone} & \multicolumn{2}{c|}{\textbf{Binary Cls}} & \multicolumn{3}{c}{\textbf{Multi-Label Cls}} \\
     & AUC $\uparrow$ & ACC $\uparrow$ & mAP $\uparrow$ & CF1 $\uparrow$ & OF1 $\uparrow$ \\
    \midrule
    \textbf{BLIP} \cite{art17} & \textbf{90.58} & \textbf{81.66} & \textbf{85.65} & \textbf{75.00} & \textbf{73.13} \\
    Siglip \cite{zhai2023sigmoid}& 89.72 & 78.77 & 83.87 & 71.77 & 69.66 \\
    Flamingo \cite{alayrac2022flamingo}& 86.60 & 75.27 & 78.34 & 64.98 & 61.77 \\
    FLAVA \cite{singh2022flava} & 82.50 & 73.70 & 73.28 & 60.83 & 55.81 \\
    \bottomrule
    \end{tabular}
    \caption{Ablation study on the choice of the Multimodal Feature Extractor. This table reports the performance of LADLE-MM when equipped with different Vision Language Models as multimodal feature extractors. Performance is evaluated using Area Under the ROC Curve (AUC) and Accuracy (ACC) for the binary classification task, alongside mean Average Precision (mAP), average per-class F1 (CF1), and average overall F1 (OF1) for the multi-label classification task. }
    \label{tab:multimodal_ablation}
\end{table*}

\subsection{Out-Of-Distribution Evaluation}
To demonstrate LADLE-MM generalization capabilities, we conduct an out-of-distribution evaluation using the VERITE~\cite{verite} benchmark.
This experiment is performed in an open-set setting, meaning that while the models are evaluated on VERITE, they have not been trained or fine-tuned on any samples from this dataset.
This evaluation not only tests the robustness of the model beyond the training distribution but also provides a clearer picture of the LADLE-MM ability to understand images and texts in a real-world scenario. Furthermore, due to the VERITE intrinsic nature, it also serves to assess the model ability to overcome unimodal bias~\cite{unimodal_bias}, as described in Section \ref{sec:experimental}.
In Table~\ref{tab:ood_results}, we present a comparative analysis against a wide range of state-of-the-art approaches, encompassing both general-purpose Large Vision-Language Models (LVLMs), such as GPT-4o and LLaVA evaluated in a zero-shot setting, and specialized misinformation detection frameworks. The table reports the Accuracy and F1 scores achieved by each method. Even though LADLE-MM is a much lighter model compared to the architectures it is benchmarked against, our approach outperforms all competing methods, achieving state-of-the-art results with an accuracy of 79.6\% and an F1 score of 83.5\%. This establishes LADLE-MM as a highly robust and effective solution for detecting multimodal misinformation in unseen domains.

\subsection{Ablation Studies}
\label{sec:ablation}
To assess the contribution of each component of our architecture, we conducted a series of ablation studies, systematically isolating specific components and training strategies. The results of all the ablation studies are summarized in Table~\ref{tab:ablation}.

\paragraph{Encoders of Feature Extraction Module}
We investigate the impact of different encoder choices in the Feature Extraction Module. Specifically we implement the following variants of LADLE-MM: (a) a ViT-Large image encoder instead of the Base one; (b) BERT language model in place of DeBERTa; (c) w/o the third multimodal branch composed of BLIP. We observe that substituting the two unimodal encoder with other variants brings to sub-optimal results compared to the full LADLE-MM configuration, but the the most critical finding emerges from the exclusion of the multimodal feature extraction branch. When the fixed pre-trained BLIP module is removed, we observe a significant degradation in performance, with binary accuracy dropping over 12 points. This strongly underlines the importance of the fixed multimodal reference space constructed with BLIP, which acts as a semantic anchor and align samples of the same class in the latent space. 
Furthermore the strategy of aggregating the representation of the last three layers prove to be more effective than just relying on the last layer, helping the model to obtain a richer representation of the input.

\begin{table}[t]
\centering
\footnotesize 
\setlength{\tabcolsep}{6pt}
\resizebox{\columnwidth}{!}{%
\begin{tabular}{llccc}
\toprule
\textbf{Encoder} & \textbf{Class} & \textbf{\shortstack{IntSim$\uparrow$}} & \textbf{\shortstack{ExtSim$\downarrow$}} & \textbf{\shortstack{SepScore$\uparrow$}} \\
\midrule
\multirow{3}{*}{\textbf{BLIP}} 
& Real & +0.5603 & +0.2752 & +0.2852 \\
& Fake & +0.3464 & +0.2752 & +0.0712 \\
\cmidrule{2-5}
& \textit{Avg} & +0.4534 & +0.2752 & \textbf{+0.1782} \\
\midrule
\multirow{3}{*}{SigLIP} 
& Real & +0.4220 & +0.2749 & +0.1471 \\
& Fake & +0.3285 & +0.2749 & +0.0536 \\
\cmidrule{2-5}
& \textit{Avg} & +0.3753 & +0.2749 & +0.1004 \\
\midrule
\multirow{3}{*}{Flamingo} 
& Real & +0.3633 & +0.1570 & +0.2063 \\
& Fake & +0.2399 & +0.1570 & +0.0829 \\
\cmidrule{2-5}
& \textit{Avg} & +0.3016 & +0.1567 & +0.1446 \\
\midrule
\multirow{3}{*}{FLAVA} 
& Real & +0.7970 & +0.7064 & +0.0906 \\
& Fake & +0.7203 & +0.7064 & +0.0139 \\
\cmidrule{2-5}
& \textit{Avg} & +0.7587 & +0.7064 & +0.0522 \\

\bottomrule
\end{tabular}%
}
\caption{Embedding space geometric separability analysis on the validation set across four multimodal encoders. Internal Similarity ($\text{IntSim}$) and External Similarity ($\text{ExtSim}$) are defined in Equations \ref{eq:int_sim} and \ref{eq:ext_sim} respectively. The Separability Score ($\text{SepScore}$) is computed as $\text{IntSim} - \text{ExtSim}$.}
\label{tab:embedding_separability}
\end{table}

\paragraph{Impact of the Multimodal Feature Extractor}
To validate our architectural design choices for the multimodal feature extraction branch, we investigated the impact of substituting our selected model with other state-of-the-art Vision-Language Models (VLMs). Specifically, we evaluated the performance of LADLE-MM using Flamingo \cite{alayrac2022flamingo}, FLAVA \cite{singh2022flava}, and Siglip \cite{zhai2023sigmoid} as alternative fixed reference spaces. The results, measured on the validation set, are summarized in Table \ref{tab:multimodal_ablation}. We observe that BLIP consistently outperforms all the other multimodal models across all key metrics. Intuitively, the superior performance of BLIP comes from its fine-grained image-text pretraining alignment, obtained with the Image-Text Contrastive Loss and Image-Text Matching Loss. 
To further validate the structural robustness of the fixed reference space, we evaluate the geometric separability of the final classification tokens across the candidate multimodal encoders.  Following \citet{teglia2025much} we compute internal similarity (IntSim) to measure the cohesion of instances belonging to the same class:
\begin{equation}
\label{eq:int_sim}
IntSim = \frac{1}{n^2 - n}  \sum_{i} \sum\limits_{j \neq i} cos(e_i, e_j), 
\end{equation}
where $e_i$ and $e_j$ represent embedding vectors of samples belonging to the same class and $n$ is the total number of samples within that class. 
Conversely, we define external similarity (ExtSim) to measure the cross-class similarity between distinct manifolds:
\begin{equation}
\label{eq:ext_sim}
\begin{aligned}
ExtSim = \frac{1}{n_A \cdot n_B} \sum_{e_i \in C_A} \sum_{e_j \in C_B} \cos(e_i, e_j)
\end{aligned}
\end{equation}
where $C_A$ and $C_B$ denote the pristine (real) and manipulated (fake) class clusters, respectively, while $n_A$ and $n_B$ represent their respective sample sizes. 
Moreover, we introduce the separation score (SepScore), defined as the difference between the internal and external similarity. A high SepScore indicates that the latent manifold tightly clusters authentic samples while establishing a clear geometric boundary against manipulated ones.
As shown in Table \ref{tab:embedding_separability}, BLIP achieves the highest separation score (+0.1782), balancing a moderate internal similarity while keeping different class embeddings well separated. In contrast, while an architecture like FLAVA exhibits high internal similarity, it suffers from severe representation collapse as showed from the high internal similarity, suggesting a tight representation space. 
This behavior demonstrates that BLIP's image-text matching pretraining creates a well-behaved fixed reference space where cross-modal anomalies are sharply distinguished, directly facilitating the downstream boundary-learning task for our Classification Module.
\begin{table}[!h]
    \centering
    \setlength{\tabcolsep}{4pt}
    \resizebox{\columnwidth}{!}{%
    \begin{tabular}{l|cc|ccc}
    \toprule
    \textbf{Configuration} & \multicolumn{2}{c|}{\textbf{Binary Cls}} & \multicolumn{3}{c}{\textbf{Multi-Label Cls}} \\
     & AUC & ACC & mAP & CF1 & OF1 \\
    \midrule
    VIT-L Backbone & 87.31 & 80.74 & 83.34 & 69.21 & 68.43\\
    BERT Backbone & 89.50 & 82.80 & 85.85 & 71.41 & 69.37\\
    w/o BLIP & 79.32 & 70.96 & 71.05 & 59.08 & 53.69 \\ 
    No Layer Avg. & 90.40 & 82.09 & 85.05 & 73.25 & 70.80 \\ 
    \midrule
    3 Fusion Blocks & 82.50 & 72.52 & 61.75 & 39.23 & 35.72 \\
    Longer Queue & 90.53 & 82.09 & 85.05 & 75.05 & 73.19 \\
    No Model Soup & 91.12 & 82.56 & 85.92 & 74.89 & 76.72 \\
    \midrule
    \textbf{LADLE-MM (full)} & \textbf{92.20} & \textbf{83.84} & \textbf{87.90} & \textbf{78.46} & \textbf{76.84} \\
    \bottomrule
    \end{tabular}%
    }
    \caption{Ablation study of different components in LADLE-MM.}
    \label{tab:ablation}
\end{table}

\paragraph{Impact of Fusion Depth and Training Strategies}
To extend our understanding of the architecture, we further investigate the impact of fusion depth and specific training dynamics. Reducing the fusion module from six to three cross-attention blocks provokes a catastrophic decline in multi-label performance, with F1 scores decreasing by nearly 40 points. This finding suggests that a deeper architecture is indispensable for capturing the fine-grained correlations required to discriminate between specific manipulation types, such as distinguishing face swaps from textual attribute alterations. 

In terms of training methodology, the model soup initialization technique reveals to be quite effective with respect to the standard training protocol, with a 1.28\% increase in terms of binary accuracy and nearly 2-point gain in mAP. Finally, increasing the length of the negative sample queue in the Manipulation-Aware Contrastive (MAC) yield minor performance benefit.

%% file: sec/6_conclusions.tex
\section{Conclusions and Future Works}

In this work, we presented LADLE-MM, a novel architecture for multimodal misinformation detection that is specifically designed to operate under limited training resources and supervision. Unlike many existing approaches, LADLE-MM does not rely on explicit spatial annotations, such as bounding boxes identifying manipulated image regions, thereby reducing annotation requirements while maintaining competitive performance.
By adopting a three-branch structure and a dual-path fusion module, LADLE-MM is able to exploit both uni-modal and multimodal features, benefiting from pre-trained models and selective fine-tuning. Our experiments on the DGM4 benchmark show that LADLE-MM achieves comparable results to the state-of-the-art models when these are trained using explicit grounding objectives and it outperform HAMMER on both binary and multi-label classification tasks when both models are trained in a limited annotation setup, i.e. without including explicit image and text grounding annotations. Although LADLE-MM does not explicitly include a grounding module, our results indicate that the model implicitly learns to focus on regions that are typically subject to manipulation, as revealed by attention visualizations contributing to the interpretability of the model. Our adoption of the model soups technique represents a novel contribution to the field, demonstrating how ensemble-inspired initialization can be effectively integrated within a multimodal framework to achieve better detection capabilities. 
Moreover, the evaluation conducted on the VERITE dataset demonstrates our model's generalization capabilities and its resistance to unimodal bias.

For future works, we plan to further reduce the dimension of the architecture to create a more lightweight model usable in real-time applications, by employing smaller pre-trained models like DistilBERT \cite{sanh2020distilbertdistilledversionbert} for text or MobileViT \cite{mehta2022mobilevit} for vision. In addition, we aim to evaluate the robustness of LADLE-MM on real-world, compressed social media content by applying compression-aware training and domain adaptation techniques.

%% file: main.bib
@String(CVPR= {IEEE Conf. Comput. Vis. Pattern Recog.})

@String(NIPS= {Adv. Neural Inform. Process. Syst.})

@String(ICME = {Int. Conf. Multimedia and Expo})

@String(ICLR = {Int. Conf. Learn. Represent.})

@String(AAAI = {AAAI})

@String(CVPRW= {IEEE Conf. Comput. Vis. Pattern Recog. Worksh.})

@String(CVPR  = {CVPR})

@String(NIPS  = {NeurIPS})

@String(ICME  =	{ICME})

@String(ICLR  = {ICLR})

@String(CVPRW= {CVPRW})

@article{fraud-ai-images-hks-review-2024,
    author = {DiResta R. and Goldstein J. A.},
    title = {How spammers and scammers leverage AI-generated images on Facebook for audience growth},
    journal = {Harvard Kennedy School (HKS) Misinformation Review},
    year = 2024
}

@article{deepfake-criminal-justice-review-2024,
	author = {Sandoval, Maria-Paz and de Almeida Vau, Maria and Solaas, John and Rodrigues, Luano},
	date = {2024/11/17},
	doi = {10.1186/s40163-024-00239-1},
	id = {Sandoval2024},
	isbn = {2193-7680},
	journal = {Crime Science},
	number = {1},
	pages = {41},
	title = {Threat of deepfakes to the criminal justice system: a systematic review},
	url = {https://doi.org/10.1186/s40163-024-00239-1},
	volume = {13},
	year = {2024}}

@article{deepfake-news-entertainment-aisoc-2025,
    author = {Lundberg E. and Mozelius P.},
    title = {The potential effects of deepfakes on news media and entertainment},
    journal = {AI \& Soc 40, 2159–2170},
    year = 2025
}

@article{vae-intro-kingma2019,
    author = {Diederik P. Kingma and Max Welling},
    title = {An Introduction to Variational Autoencoders},
    journal = {Foundations and Trends in Machine Learning},
    year = 2019
}

@INPROCEEDINGS{diffusion-ldm-rombach2021,
  author={Rombach, Robin and Blattmann, Andreas and Lorenz, Dominik and Esser, Patrick and Ommer, Björn},
  booktitle={2022 IEEE/CVF Conference on Computer Vision and Pattern Recognition (CVPR)}, 
  title={High-Resolution Image Synthesis with Latent Diffusion Models}, 
  year={2022},
  volume={},
  number={},
  pages={10674-10685},
  doi={10.1109/CVPR52688.2022.01042}}

@article{gan-goodfellow2014,
author = {Goodfellow, Ian and Pouget-Abadie, Jean and Mirza, Mehdi and Xu, Bing and Warde-Farley, David and Ozair, Sherjil and Courville, Aaron and Bengio, Yoshua},
title = {Generative adversarial networks},
year = {2020},
issue_date = {November 2020},
publisher = {Association for Computing Machinery},
address = {New York, NY, USA},
volume = {63},
number = {11},
issn = {0001-0782},
url = {https://doi.org/10.1145/3422622},
doi = {10.1145/3422622},
journal = {Commun. ACM},
month = oct,
pages = {139–144},
numpages = {6}
}

@article{llm-textgen-overview-2024,
  author={Bengesi, Staphord and El-Sayed, Hoda and Sarker, MD Kamruzzaman and Houkpati, Yao and Irungu, John and Oladunni, Timothy},
  journal={IEEE Access}, 
  title={Advancements in Generative AI: A Comprehensive Review of GANs, GPT, Autoencoders, Diffusion Model, and Transformers}, 
  year={2024},
  volume={12},
  number={},
  pages={69812-69837},
  doi={10.1109/ACCESS.2024.3397775}}

@inproceedings{dgm4,
    title={Detecting and Grounding Multi-Modal Media Manipulation},
    author={Shao, Rui and Wu, Tianxing and Liu, Ziwei},
    booktitle={IEEE Conference on Computer Vision and Pattern Recognition (CVPR)},
    year={2023}
}

@article{viki,
author = {Li, Qilei and Gao, Mingliang and Zhang, Guisheng and Zhai, Wenzhe and Chen, Jinyong and Jeon, Gwanggil},
title = {Towards Multimodal Disinformation Detection by Vision-language Knowledge Interaction},
year = {2024},
issue_date = {Feb 2024},
publisher = {Elsevier Science Publishers B. V.},
address = {NLD},
volume = {102},
number = {C},
issn = {1566-2535},
url = {https://doi.org/10.1016/j.inffus.2023.102037},
doi = {10.1016/j.inffus.2023.102037},
journal = {Inf. Fusion},
month = feb,
numpages = {11}
}

@inproceedings{ttblip,
    author = {Eunjee Choi and Jong-Kook Kim},
    title = {TT-BLIP: Enhancing Fake News Detection Using BLIP and Tri-Transformer},
    booktitle = {27th International Conference on Information Fusion, FUSION 2024},
    year = {2024}
}

@article{exposing_deception,
  title={Exposing the Deception: Uncovering More Forgery Clues for Deepfake Detection},
  author={Ba, Zhongjie and Liu, Qingyu and Liu, Zhenguang and Wu, Shuang and Lin, Feng and Lu, Li and Ren, Kui},
  journal={arXiv preprint arXiv:2403.01786},
  year={2024}
}

@inproceedings{art12,
author = {Tan, Chuangchuang and Zhao, Yao and Wei, Shikui and Gu, Guanghua and Liu, Ping and Wei, Yunchao},
title = {Frequency-aware deepfake detection: improving generalizability through frequency space domain learning},
year = {2024},
isbn = {978-1-57735-887-9},
publisher = {AAAI Press},
url = {https://doi.org/10.1609/aaai.v38i5.28310},
doi = {10.1609/aaai.v38i5.28310},
booktitle = {Proceedings of the Thirty-Eighth AAAI Conference on Artificial Intelligence and Thirty-Sixth Conference on Innovative Applications of Artificial Intelligence and Fourteenth Symposium on Educational Advances in Artificial Intelligence},
articleno = {562},
numpages = {9},
series = {AAAI'24/IAAI'24/EAAI'24}
}

@article{art15,
    author = "Dosovitskiy, Alexey and Beyer, Lucas and Kolsnikov, Alexander and Weissenborn, Dirk and Zhai, Xiaohua and Unterthiner, Thomas and Dehghani, Mostafa and Minderer, Matthias and Heigold, Georg and Gelly, Sylvain and others",
    journal = "arXiv preprint arXiv:2010.11929",
    title = "{An Image is Worth 16x16 Words: Transformers for Image Recognition at Scale}",
    year = "2020"
}

@inproceedings{
art16,
title={DEBERTA: DECODING-ENHANCED BERT WITH DISENTANGLED ATTENTION},
author={Pengcheng He and Xiaodong Liu and Jianfeng Gao and Weizhu Chen},
booktitle={International Conference on Learning Representations},
year={2021},
url={https://openreview.net/forum?id=XPZIaotutsD}
}

@inproceedings{art17,
      title={BLIP: Bootstrapping Language-Image Pre-training for Unified Vision-Language Understanding and Generation}, 
      author={Junnan Li and Dongxu Li and Caiming Xiong and Steven Hoi},
      year={2022},
      booktitle={ICML},
}

@inproceedings{visual_news,
    title = "Visual News: Benchmark and Challenges in News Image Captioning",
    author = "Liu, Fuxiao  and
      Wang, Yinghan  and
      Wang, Tianlu  and
      Ordonez, Vicente",
    editor = "Moens, Marie-Francine  and
      Huang, Xuanjing  and
      Specia, Lucia  and
      Yih, Scott Wen-tau",
    booktitle = "Proceedings of the 2021 Conference on Empirical Methods in Natural Language Processing",
    month = nov,
    year = "2021",
    address = "Online and Punta Cana, Dominican Republic",
    publisher = "Association for Computational Linguistics",
    url = "https://aclanthology.org/2021.emnlp-main.542/",
    doi = "10.18653/v1/2021.emnlp-main.542",
    pages = "6761--6771",
}

@article{vlp-gf,
	author = {Zhang, Guisheng and Gao, Mingliang and Li, Qilei and Zhai, Wenzhe and Jeon, Gwanggil},
	date = {2024/11/01},
	doi = {10.1007/s12559-024-10316-x},
	id = {Zhang2024},
	isbn = {1866-9964},
	journal = {Cognitive Computation},
	number = {6},
	pages = {2953--2966},
	title = {Multi-Modal Generative DeepFake Detection via Visual-Language Pretraining with Gate Fusion for Cognitive Computation},
	url = {https://doi.org/10.1007/s12559-024-10316-x},
	volume = {16},
	year = {2024}}

@InProceedings{beyond_attention,
    author    = {Chefer, Hila and Gur, Shir and Wolf, Lior},
    title     = {Transformer Interpretability Beyond Attention Visualization},
    booktitle = {Proceedings of the IEEE/CVF Conference on Computer Vision and Pattern Recognition (CVPR)},
    month     = {June},
    year      = {2021},
    pages     = {782-791}
}

@InProceedings{art21,
  title = 	 {ViLT: Vision-and-Language Transformer Without Convolution or Region Supervision},
  author =       {Kim, Wonjae and Son, Bokyung and Kim, Ildoo},
  booktitle = 	 {Proceedings of the 38th International Conference on Machine Learning},
  pages = 	 {5583--5594},
  year = 	 {2021},
  editor = 	 {Meila, Marina and Zhang, Tong},
  volume = 	 {139},
  series = 	 {Proceedings of Machine Learning Research},
  month = 	 {18--24 Jul},
  publisher =    {PMLR},
  url = 	 {https://proceedings.mlr.press/v139/kim21k.html},
}

@InProceedings{art22,
  title = 	 {ViLT: Vision-and-Language Transformer Without Convolution or Region Supervision},
  author =       {Kim, Wonjae and Son, Bokyung and Kim, Ildoo},
  booktitle = 	 {Proceedings of the 38th International Conference on Machine Learning},
  pages = 	 {5583--5594},
  year = 	 {2021},
  editor = 	 {Meila, Marina and Zhang, Tong},
  volume = 	 {139},
  series = 	 {Proceedings of Machine Learning Research},
  month = 	 {18--24 Jul},
  publisher =    {PMLR},
  url = 	 {https://proceedings.mlr.press/v139/kim21k.html},
}

@Inbook{art23,
author="Kurita, Takio",
title="Principal Component Analysis (PCA)",
bookTitle="Computer Vision: A Reference Guide",
year="2019",
publisher="Springer International Publishing",
address="Cham",
pages="1--4",
isbn="978-3-030-03243-2",
doi="10.1007/978-3-030-03243-2_649-1",
url="https://doi.org/10.1007/978-3-030-03243-2_649-1"
}

@article{art24,
	author = {Zhang, Guisheng and Gao, Mingliang and Li, Qilei and Zhai, Wenzhe and Jeon, Gwanggil},
	date = {2024/11/01},
	doi = {10.1007/s12559-024-10316-x},
	id = {Zhang2024},
	isbn = {1866-9964},
	journal = {Cognitive Computation},
	number = {6},
	pages = {2953--2966},
	title = {Multi-Modal Generative DeepFake Detection via Visual-Language Pretraining with Gate Fusion for Cognitive Computation},
	url = {https://doi.org/10.1007/s12559-024-10316-x},
	volume = {16},
	year = {2024}}

@InProceedings{t2i-dalle-ramesh2021,
  title = 	 {Zero-Shot Text-to-Image Generation},
  author =       {Ramesh, Aditya and Pavlov, Mikhail and Goh, Gabriel and Gray, Scott and Voss, Chelsea and Radford, Alec and Chen, Mark and Sutskever, Ilya},
  booktitle = 	 {Proceedings of the 38th International Conference on Machine Learning},
  pages = 	 {8821--8831},
  year = 	 {2021},
  editor = 	 {Meila, Marina and Zhang, Tong},
  volume = 	 {139},
  series = 	 {Proceedings of Machine Learning Research},
  month = 	 {18--24 Jul},
  publisher =    {PMLR},
  url = 	 {https://proceedings.mlr.press/v139/ramesh21a.html}
}

@inproceedings{captioning-show-and-tell-vinyals2015,
  author = {Vinyals, Oriol and Toshev, Alexander and Bengio, Samy and Erhan, Dumitru},
  booktitle = {CVPR},
  ee = {https://www.wikidata.org/entity/Q62763559},
  isbn = {978-1-4673-6964-0},
  pages = {3156-3164},
  publisher = {IEEE Computer Society},
  title = {Show and tell: A neural image caption generator.},
  url = {http://dblp.uni-trier.de/db/conf/cvpr/cvpr2015.html#VinyalsTBE15},
  year = 2015
}

@inproceedings{captioning-fast-multi-length-2023,
author = {Hu, Jia Cheng and Cavicchioli, Roberto and Capotondi, Alessandro},
year = {2023},
month = {12},
pages = {2173-2182},
title = {Exploiting Multiple Sequence Lengths in Fast End to End Training for Image Captioning},
doi = {10.1109/BigData59044.2023.10386812},
booktitle={}
}

@ARTICLE{hammerpp,
  author={Shao, Rui and Wu, Tianxing and Wu, Jianlong and Nie, Liqiang and Liu, Ziwei},
  journal={IEEE Transactions on Pattern Analysis and Machine Intelligence}, 
  title={Detecting and Grounding Multi-Modal Media Manipulation and Beyond}, 
  year={2024},
  volume={46},
  number={8},
  pages={5556-5574},
  doi={10.1109/TPAMI.2024.3367749}}

@InProceedings{model_soup,
  title = 	 {Model soups: averaging weights of multiple fine-tuned models improves accuracy without increasing inference time},
  author =       {Wortsman, Mitchell and Ilharco, Gabriel and Gadre, Samir Ya and Roelofs, Rebecca and Gontijo-Lopes, Raphael and Morcos, Ari S and Namkoong, Hongseok and Farhadi, Ali and Carmon, Yair and Kornblith, Simon and Schmidt, Ludwig},
  booktitle = 	 {Proceedings of the 39th International Conference on Machine Learning},
  pages = 	 {23965--23998},
  year = 	 {2022},
  editor = 	 {Chaudhuri, Kamalika and Jegelka, Stefanie and Song, Le and Szepesvari, Csaba and Niu, Gang and Sabato, Sivan},
  volume = 	 {162},
  series = 	 {Proceedings of Machine Learning Research},
  month = 	 {17--23 Jul},
  publisher =    {PMLR},
  url = 	 {https://proceedings.mlr.press/v162/wortsman22a.html}
}

@inproceedings{mm-fakenews-contrastive-wang2023,
  title={Cross-modal contrastive learning for multimodal fake news detection},
  author={Wang, Longzheng and Zhang, Chuang and Xu, Hongbo and Xu, Yongxiu and Xu, Xiaohan and Wang, Siqi},
  booktitle={Proceedings of the 31st ACM international conference on multimedia},
  pages={5696--5704},
  year={2023}
}

@Article{mm-deepfake-detection-salvi2023,
AUTHOR = {Salvi, Davide and Liu, Honggu and Mandelli, Sara and Bestagini, Paolo and Zhou, Wenbo and Zhang, Weiming and Tubaro, Stefano},
TITLE = {A Robust Approach to Multimodal Deepfake Detection},
JOURNAL = {Journal of Imaging},
VOLUME = {9},
YEAR = {2023},
NUMBER = {6},
ARTICLE-NUMBER = {122},
URL = {https://www.mdpi.com/2313-433X/9/6/122},
PubMedID = {37367470},
ISSN = {2313-433X},
DOI = {10.3390/jimaging9060122}
}

@article{color_components,
  title={Identification of deep network generated images using disparities in color components},
  author={Li, Haodong and Li, Bin and Tan, Shunquan and Huang, Jiwu},
  journal={Signal Processing},
  volume={174},
  pages={107616},
  year={2020},
  publisher={Elsevier}
}

@article{detecting_gan,
  title={Detecting GAN-generated Imagery using Color Cues},
  author={Scott McCloskey and Michael Albright},
  journal={ArXiv},
  year={2018},
  volume={abs/1812.08247},
  url={https://api.semanticscholar.org/CorpusID:56517594}
}

@article{lightning_inconsistencies,
  title={Lighting (in) consistency of paint by text},
  author={Farid, Hany},
  journal={arXiv preprint arXiv:2207.13744},
  year={2022}
}

@article{perspective_inconsistency,
  title={Perspective (in) consistency of paint by text},
  author={Farid, Hany},
  journal={arXiv preprint arXiv:2206.14617},
  year={2022}
}

@inproceedings{afchar2018mesonet,
  title={Mesonet: a compact facial video forgery detection network},
  author={Afchar, Darius and Nozick, Vincent and Yamagishi, Junichi and Echizen, Isao},
  booktitle={2018 IEEE international workshop on information forensics and security (WIFS)},
  pages={1--7},
  year={2018},
  organization={IEEE}
}

@inproceedings{liu2020global,
  title={Global texture enhancement for fake face detection in the wild},
  author={Liu, Zhengzhe and Qi, Xiaojuan and Torr, Philip HS},
  booktitle={Proceedings of the IEEE/CVF conference on computer vision and pattern recognition},
  pages={8060--8069},
  year={2020}
}

@inproceedings{co_occurrency_detection,
    author = {Lakshmanan Nataraj and Tajuddin Manhar Mohammed and B. S. Manjunath and Shivkumar Chandrasekaran and Arjuna Flenner and Jawadul H. Bappy and Amit K. Roy-Chowdhury},
    title = {Detecting GAN generated Fake Images using Co-occurrence Matrices},
    booktitle = {Proceedings of Society for Imaging Science and Technology. Symp. on Electronic Imaging: Media Watermarking, Security, and Forensics, 2019, pp 532-1 - 532-7},
    year = {2019}
}

@article{zhong2023patchcraft,
  title={Patchcraft: Exploring texture patch for efficient ai-generated image detection},
  author={Zhong, Nan and Xu, Yiran and Li, Sheng and Qian, Zhenxing and Zhang, Xinpeng},
  journal={arXiv preprint arXiv:2311.12397},
  year={2023}
}

@inproceedings{tan2023learning,
  title={Learning on gradients: Generalized artifacts representation for gan-generated images detection},
  author={Tan, Chuangchuang and Zhao, Yao and Wei, Shikui and Gu, Guanghua and Wei, Yunchao},
  booktitle={Proceedings of the IEEE/CVF Conference on Computer Vision and Pattern Recognition},
  pages={12105--12114},
  year={2023}
}

@inproceedings{marra2019gans,
  title={Do gans leave artificial fingerprints?},
  author={Marra, Francesco and Gragnaniello, Diego and Verdoliva, Luisa and Poggi, Giovanni},
  booktitle={2019 IEEE conference on multimedia information processing and retrieval (MIPR)},
  pages={506--511},
  year={2019},
  organization={IEEE}
}

@inproceedings{yu2019attributing,
  title={Attributing fake images to gans: Learning and analyzing gan fingerprints},
  author={Yu, Ning and Davis, Larry S and Fritz, Mario},
  booktitle={Proceedings of the IEEE/CVF international conference on computer vision},
  pages={7556--7566},
  year={2019}
}

@inproceedings{durall2020watch,
  title={Watch your up-convolution: Cnn based generative deep neural networks are failing to reproduce spectral distributions},
  author={Durall, Ricard and Keuper, Margret and Keuper, Janis},
  booktitle={Proceedings of the IEEE/CVF conference on computer vision and pattern recognition},
  pages={7890--7899},
  year={2020}
}

@INPROCEEDINGS{10156981,
  author={Papa, Lorenzo and Faiella, Lorenzo and Corvitto, Luca and Maiano, Luca and Amerini, Irene},
  booktitle={2023 11th International Workshop on Biometrics and Forensics (IWBF)}, 
  title={On the use of Stable Diffusion for creating realistic faces: from generation to detection}, 
  year={2023},
  volume={},
  number={},
  pages={1-6},
  doi={10.1109/IWBF57495.2023.10156981}}

@article{dzanic2020fourier,
  title={Fourier spectrum discrepancies in deep network generated images},
  author={Dzanic, Tarik and Shah, Karan and Witherden, Freddie},
  journal={Advances in neural information processing systems},
  volume={33},
  pages={3022--3032},
  year={2020}
}

@article{liu2023survey,
  title={A survey of visual transformers},
  author={Liu, Yang and Zhang, Yao and Wang, Yixin and Hou, Feng and Yuan, Jin and Tian, Jiang and Zhang, Yang and Shi, Zhongchao and Fan, Jianping and He, Zhiqiang},
  journal={IEEE Transactions on Neural Networks and Learning Systems},
  year={2023},
  publisher={IEEE}
}

@inproceedings{dong2022protecting,
  title={Protecting celebrities from deepfake with identity consistency transformer},
  author={Dong, Xiaoyi and Bao, Jianmin and Chen, Dongdong and Zhang, Ting and Zhang, Weiming and Yu, Nenghai and Chen, Dong and Wen, Fang and Guo, Baining},
  booktitle={Proceedings of the IEEE/CVF Conference on Computer Vision and Pattern Recognition},
  pages={9468--9478},
  year={2022}
}

@inproceedings{zhuang2022uia,
  title={UIA-ViT: Unsupervised inconsistency-aware method based on vision transformer for face forgery detection},
  author={Zhuang, Wanyi and Chu, Qi and Tan, Zhentao and Liu, Qiankun and Yuan, Haojie and Miao, Changtao and Luo, Zixiang and Yu, Nenghai},
  booktitle={European conference on computer vision},
  pages={391--407},
  year={2022},
  organization={Springer}
}

@article{heo2021deepfake,
  title={Deepfake detection scheme based on vision transformer and distillation},
  author={Heo, Young-Jin and Choi, Young-Ju and Lee, Young-Woon and Kim, Byung-Gyu},
  journal={arXiv preprint arXiv:2104.01353},
  year={2021}
}

@article{wang2023deep,
  title={Deep convolutional pooling transformer for deepfake detection},
  author={Wang, Tianyi and Cheng, Harry and Chow, Kam Pui and Nie, Liqiang},
  journal={ACM transactions on multimedia computing, communications and applications},
  volume={19},
  number={6},
  pages={1--20},
  year={2023},
  publisher={ACM New York, NY}
}

@inproceedings{TheC3,
  title={The Claude 3 Model Family: Opus, Sonnet, Haiku},
  author={Anthropic},
  url={https://api.semanticscholar.org/CorpusID:268232499},
  year={2024},
  booktitle={The Claude 3 Model Family: Opus, Sonnet, Haiku}
}

@misc{openai2023gpt4,
  author       = {OpenAI},
  title        = {GPT-4 Technical Report},
  year         = {2023},
  url          = {https://arxiv.org/abs/2303.08774},
}

@article{gemini,
author = {Gemini Team},
year = {2023},
month = {12},
pages = {},
title = {Gemini: A Family of Highly Capable Multimodal Models},
doi = {10.48550/arXiv.2312.11805}
}

@article{art61,
author = {Al-khazraji, Samer and Saleh, Hassan and KHALID, Adil and Mishkhal, Israa},
year = {2023},
month = {10},
pages = {429-441},
title = {Impact of Deepfake Technology on Social Media: Detection, Misinformation and Societal Implications},
volume = {23},
journal = {The Eurasia Proceedings of Science Technology Engineering and Mathematics},
doi = {10.55549/epstem.1371792}
}

@article{Atlam2025SLMDFS,
  title        = {SLM-DFS: A systematic literature map of deepfake spread on social media},
  author       = {El‐Sayed Atlam and Malik Almaliki and Ghada Elmarhomy and Abdulqader M.~Almars and Awatif M.A.~Elsiddieg and Rasha ElAgamy},
  journal      = {Alexandria Engineering Journal},
  volume       = {111},
  pages        = {446--455},
  year         = {2025},
  doi          = {10.1016/j.aej.2024.10.076},
}

@article{LundbergEbba,
author = {Lundberg, Ebba and Mozelius, Peter},
year = {2024},
month = {10},
pages = {2159-2170},
title = {The potential effects of deepfakes on news media and entertainment},
volume = {40},
journal = {AI and SOCIETY},
doi = {10.1007/s00146-024-02072-1}
}

@misc{grok,
  author = {{xAI}},
  title = {Grok-3},
  year = {2025},
  url = {https://x.ai},
}

@inproceedings{10.1145/3219819.3219903,
author = {Wang, Yaqing and Ma, Fenglong and Jin, Zhiwei and Yuan, Ye and Xun, Guangxu and Jha, Kishlay and Su, Lu and Gao, Jing},
title = {EANN: Event Adversarial Neural Networks for Multi-Modal Fake News Detection},
year = {2018},
isbn = {9781450355520},
publisher = {Association for Computing Machinery},
address = {New York, NY, USA},
url = {https://doi.org/10.1145/3219819.3219903},
doi = {10.1145/3219819.3219903},
booktitle = {Proceedings of the 24th ACM SIGKDD International Conference on Knowledge Discovery \& Data Mining},
pages = {849–857},
numpages = {9},
location = {London, United Kingdom},
series = {KDD '18}
}

@inproceedings{10.1145/3308558.3313552,
author = {Khattar, Dhruv and Goud, Jaipal Singh and Gupta, Manish and Varma, Vasudeva},
title = {MVAE: Multimodal Variational Autoencoder for Fake News Detection},
year = {2019},
isbn = {9781450366748},
publisher = {Association for Computing Machinery},
address = {New York, NY, USA},
url = {https://doi.org/10.1145/3308558.3313552},
doi = {10.1145/3308558.3313552},
booktitle = {The World Wide Web Conference},
pages = {2915–2921},
numpages = {7},
location = {San Francisco, CA, USA},
series = {WWW '19}
}

@inproceedings{vaswani,
author = {Vaswani, Ashish and Shazeer, Noam and Parmar, Niki and Uszkoreit, Jakob and Jones, Llion and Gomez, Aidan N. and Kaiser, \L{}ukasz and Polosukhin, Illia},
title = {Attention is all you need},
year = {2017},
isbn = {9781510860964},
publisher = {Curran Associates Inc.},
address = {Red Hook, NY, USA},
booktitle = {Proceedings of the 31st International Conference on Neural Information Processing Systems},
pages = {6000–6010},
numpages = {11},
location = {Long Beach, California, USA},
series = {NIPS'17}
}

@INPROCEEDINGS{spotfake,
  author={Singhal, Shivangi and Shah, Rajiv Ratn and Chakraborty, Tanmoy and Kumaraguru, Ponnurangam and Satoh, Shin'ichi},
  booktitle={2019 IEEE Fifth International Conference on Multimedia Big Data (BigMM)}, 
  title={SpotFake: A Multi-modal Framework for Fake News Detection}, 
  year={2019},
  volume={},
  number={},
  pages={39-47},
  doi={10.1109/BigMM.2019.00-44}}

@inproceedings{bert,
    title = "{BERT}: Pre-training of Deep Bidirectional Transformers for Language Understanding",
    author = "Devlin, Jacob  and
      Chang, Ming-Wei  and
      Lee, Kenton  and
      Toutanova, Kristina",
    editor = "Burstein, Jill  and
      Doran, Christy  and
      Solorio, Thamar",
    booktitle = "Proceedings of the 2019 Conference of the North {A}merican Chapter of the Association for Computational Linguistics: Human Language Technologies, Volume 1 (Long and Short Papers)",
    month = jun,
    year = "2019",
    address = "Minneapolis, Minnesota",
    publisher = "Association for Computational Linguistics",
    url = "https://aclanthology.org/N19-1423/",
    doi = "10.18653/v1/N19-1423",
    pages = "4171--4186",
}

@inproceedings{vgg,
  author = {Simonyan, Karen and Zisserman, Andrew},
  booktitle = {ICLR},
  editor = {Bengio, Yoshua and LeCun, Yann},
  ee = {http://arxiv.org/abs/1409.1556},
  title = {Very Deep Convolutional Networks for Large-Scale Image Recognition.},
  url = {http://dblp.uni-trier.de/db/conf/iclr/iclr2015.html#SimonyanZ14a},
  year = 2015
}

@INPROCEEDINGS{jia2024chatgpt,
  author={Jia, Shan and Lyu, Reilin and Zhao, Kangran and Chen, Yize and Yan, Zhiyuan and Ju, Yan and Hu, Chuanbo and Li, Xin and Wu, Baoyuan and Lyu, Siwei},
  booktitle={2024 IEEE/CVF Conference on Computer Vision and Pattern Recognition Workshops (CVPRW)}, 
  title={Can ChatGPT Detect DeepFakes? A Study of Using Multimodal Large Language Models for Media Forensics}, 
  year={2024},
  volume={},
  number={},
  pages={4324-4333},
  doi={10.1109/CVPRW63382.2024.00436}}

@inproceedings{liu2024fka,
    title={FKA-Owl: Advancing Multimodal Fake News Detection through Knowledge-Augmented LVLMs},
    author={Liu, Xuannan and Li, Peipei and Huang, Huaibo and Li, Zekun and Cui, Xing and Liang, Jiahao and Qin, Lixiong and Deng, Weihong and He, Zhaofeng},
    booktitle={ACM MM},
    year={2024}
}

@INPROCEEDINGS{msca,
  author={Huang, Zhongqiang and Hu, Yuxue and Zeng, Zhi and Li, Xiang and Sha, Ying},
  booktitle={2023 IEEE International Conference on Multimedia and Expo (ICME)}, 
  title={Multimodal Stacked Cross Attention Network for Fine-Grained Fake News Detection}, 
  year={2023},
  volume={},
  number={},
  pages={2837-2842},
  doi={10.1109/ICME55011.2023.00482}
}

@article{Oord2018RepresentationLW,
  title={Representation Learning with Contrastive Predictive Coding},
  author={A{\"a}ron van den Oord and Yazhe Li and Oriol Vinyals},
  journal={ArXiv},
  year={2018},
  volume={abs/1807.03748},
  url={https://api.semanticscholar.org/CorpusID:49670925}
}

@misc{chatgpt4o,
  author = {{OpenAI}},
  title = {ChatGPT-4o},
  year = {2024},
  url = {https://openai.com/index/hello-gpt-4o/},
}

@inproceedings{qi2024sniffer,
  title={Sniffer: Multimodal large language model for explainable out-of-context misinformation detection},
  author={Qi, Peng and Yan, Zehong and Hsu, Wynne and Lee, Mong Li},
  booktitle={Proceedings of the IEEE/CVF conference on computer vision and pattern recognition},
  pages={13052--13062},
  year={2024}
}

@InProceedings{tahmabi,
  author    = {Sahar Tahmasebi, Eric Müller-Budack and Ralph Ewerth},
  booktitle = {ACM International Conference on Information and Knowledge Management, CIKM 2024, Boise, Idaho, United States of America, October 21-25, 2024},
  title     = {Multimodal Misinformation Detection using Large Vision-Language Models},
  year      = {2024},
  doi       = {10.1145/3627673.3679826},
}

@article{frequency,
author = {Liu, Huan and Tan, Zichang and Chen, Qiang and Wei, Yunchao and Zhao, Yao and Wang, Jingdong},
title = {Unified Frequency-Assisted Transformer Framework for Detecting and Grounding Multi-modal Manipulation},
year = {2024},
issue_date = {Mar 2025},
publisher = {Kluwer Academic Publishers},
address = {USA},
volume = {133},
number = {3},
issn = {0920-5691},
url = {https://doi.org/10.1007/s11263-024-02245-x},
doi = {10.1007/s11263-024-02245-x},
journal = {Int. J. Comput. Vision},
month = oct,
pages = {1392–1409},
numpages = {18}
}

@misc{moco_loss,
  author = {He, Kaiming and Fan, Haoqi and Wu, Yuxin and Xie, Saining and Girshick, Ross},
  description = {Momentum Contrast for Unsupervised Visual Representation Learning},
  note = {cite arxiv:1911.05722Comment: CVPR 2020 camera-ready. Code:  https://github.com/facebookresearch/moco},
  title = {Momentum Contrast for Unsupervised Visual Representation Learning},
  url = {http://arxiv.org/abs/1911.05722},
  year = 2019
}

@inproceedings{sanh2020distilbertdistilledversionbert,
  title={DistilBERT, a distilled version of BERT: smaller, faster, cheaper and lighter},
  author={Sanh, Victor and Debut, Lysandre and Chaumond, Julien and Wolf, Thomas},
  booktitle={5th Edition of the EMC2 Workshop (Energy Efficient Machine Learning and Cognitive Computing) at NeurIPS 2019},
  year={2019}
}

@inproceedings{
mehta2022mobilevit,
title={MobileViT: Light-weight, General-purpose, and Mobile-friendly Vision Transformer},
author={Sachin Mehta and Mohammad Rastegari},
booktitle={International Conference on Learning Representations},
year={2022},
url={https://openreview.net/forum?id=vh-0sUt8HlG}
}

@inproceedings{unimodal_bias,
author = {Zhang, Yedi and Latham, Peter E. and Saxe, Andrew},
title = {Understanding unimodal bias in multimodal deep linear networks},
year = {2024},
publisher = {JMLR.org},
booktitle = {Proceedings of the 41st International Conference on Machine Learning},
articleno = {2441},
numpages = {26},
location = {Vienna, Austria},
series = {ICML'24}
}

@Article{verite,
author={Papadopoulos, Stefanos-Iordanis
and Koutlis, Christos
and Papadopoulos, Symeon
and Petrantonakis, Panagiotis C.},
title={VERITE: a Robust benchmark for multimodal misinformation detection accounting for unimodal bias},
journal={International Journal of Multimedia Information Retrieval},
year={2024},
month={Jan},
day={08},
volume={13},
number={1},
pages={4},
issn={2192-662X},
doi={10.1007/s13735-023-00312-6},
url={https://doi.org/10.1007/s13735-023-00312-6}
}

@inproceedings{profile_spreading,
author = {Shu, Kai and Zhou, Xinyi and Wang, Suhang and Zafarani, Reza and Liu, Huan},
title = {The role of user profiles for fake news detection},
year = {2020},
isbn = {9781450368681},
publisher = {Association for Computing Machinery},
address = {New York, NY, USA},
url = {https://doi.org/10.1145/3341161.3342927},
doi = {10.1145/3341161.3342927},
booktitle = {Proceedings of the 2019 IEEE/ACM International Conference on Advances in Social Networks Analysis and Mining},
pages = {436–439},
numpages = {4},
location = {Vancouver, British Columbia, Canada},
series = {ASONAM '19}
}

@article{profile_spreading_2,
title = {Enhancing fake news detection through estimating user tendencies to spread fake news},
journal = {Data and Information Management},
volume = {10},
number = {2},
pages = {100115},
year = {2026},
issn = {2543-9251},
doi = {https://doi.org/10.1016/j.dim.2025.100115},
url = {https://www.sciencedirect.com/science/article/pii/S2543925125000233},
author = {Ahmad Hashemi and Mohammad Reza Moosavi and Wei Shi and Anastasia Giachanou},
}

@article{social_network_spread,
author = {Sivasankari, S. and Vadivu, G.},
title = {Tracing the fake news propagation path using social network analysis},
year = {2022},
issue_date = {Dec 2022},
publisher = {Springer-Verlag},
address = {Berlin, Heidelberg},
volume = {26},
number = {23},
issn = {1432-7643},
url = {https://doi.org/10.1007/s00500-021-06043-2},
doi = {10.1007/s00500-021-06043-2},
journal = {Soft Comput.},
month = dec,
pages = {12883–12891},
numpages = {9}
}

@article{svm_fake_news,
title = {Detection of fake news from social media using support vector machine learning algorithms},
journal = {Measurement: Sensors},
volume = {32},
pages = {101028},
year = {2024},
issn = {2665-9174},
doi = {https://doi.org/10.1016/j.measen.2024.101028},
url = {https://www.sciencedirect.com/science/article/pii/S2665917424000047},
author = {M. Sudhakar and K.P. Kaliyamurthie},
}

@INPROCEEDINGS{bow_fake_news,
  author={Kumar Jain, Mayank and Gopalani, Dinesh and Kumar Meena, Yogesh and Kumar, Rajesh},
  booktitle={2020 IEEE 7th Uttar Pradesh Section International Conference on Electrical, Electronics and Computer Engineering (UPCON)}, 
  title={Machine Learning based Fake News Detection using linguistic features and word vector features}, 
  year={2020},
  volume={},
  number={},
  pages={1-6},
  doi={10.1109/UPCON50219.2020.9376576}}

@article{pos_fake_news,
  author={Afreen Kansal},
  title={Fake News Detection Using Pos Tagging and Machine Learning},
  journal={Journal of Applied Security Research},
  volume={18},
  number={2},
  pages={164--179},
  year={2023},
  note={Published online: 01 Sep 2021},
  doi={10.1080/19361610.2021.1963605},
  url={https://doi.org/10.1080/19361610.2021.1963605},
  publisher={Taylor \& Francis}
}

@inproceedings{ngram_fake_news,
author = {Wynne, Hnin Ei and Wint, Zar Zar},
title = {Content Based Fake News Detection Using N-Gram Models},
year = {2020},
isbn = {9781450371797},
publisher = {Association for Computing Machinery},
address = {New York, NY, USA},
url = {https://doi.org/10.1145/3366030.3366116},
doi = {10.1145/3366030.3366116},
booktitle = {Proceedings of the 21st International Conference on Information Integration and Web-Based Applications \& Services},
pages = {669–673},
numpages = {5},
location = {Munich, Germany},
series = {iiWAS2019}
}

@INPROCEEDINGS{bi-lstm-fake-news,
  author={Mahara, Govind Singh and Gangele, Sharad},
  booktitle={2022 IEEE 1st International Conference on Data, Decision and Systems (ICDDS)}, 
  title={Fake news detection: A RNN-LSTM, Bi-LSTM based deep learning approach}, 
  year={2022},
  volume={},
  number={},
  pages={01-06},
  doi={10.1109/ICDDS56399.2022.10037403}}

@article{bi-lstm-fake-news-2,
title = {Fake News Detection using Bi-directional LSTM-Recurrent Neural Network},
journal = {Procedia Computer Science},
volume = {165},
pages = {74-82},
year = {2019},
note = {2nd International Conference on Recent Trends in Advanced Computing ICRTAC -DISRUP - TIV INNOVATION , 2019 November 11-12, 2019},
issn = {1877-0509},
doi = {https://doi.org/10.1016/j.procs.2020.01.072},
url = {https://www.sciencedirect.com/science/article/pii/S1877050920300806},
author = {Pritika Bahad and Preeti Saxena and Raj Kamal},
}

@article{bert-fake-news-1,
title = {Fake News Detection using Bi-directional LSTM-Recurrent Neural Network},
journal = {Procedia Computer Science},
volume = {165},
pages = {74-82},
year = {2019},
note = {2nd International Conference on Recent Trends in Advanced Computing ICRTAC -DISRUP - TIV INNOVATION , 2019 November 11-12, 2019},
issn = {1877-0509},
doi = {https://doi.org/10.1016/j.procs.2020.01.072},
url = {https://www.sciencedirect.com/science/article/pii/S1877050920300806},
author = {Pritika Bahad and Preeti Saxena and Raj Kamal},
}

@article{bert-fake-news-2,
   title={Fake news detection using dual BERT deep neural networks},
   volume={83},
   ISSN={1573-7721},
   url={http://dx.doi.org/10.1007/s11042-023-17115-w},
   DOI={10.1007/s11042-023-17115-w},
   number={15},
   journal={Multimedia Tools and Applications},
   publisher={Springer Science and Business Media LLC},
   author={Farokhian, Mahmood and Rafe, Vahid and Veisi, Hadi},
   year={2023},
   month=oct, pages={43831–43848} }

@article{bert-fake-news-3,
title = {Fake News Detection Based on BERT Multi-domain and Multi-modal Fusion Network},
journal = {Computer Vision and Image Understanding},
volume = {252},
pages = {104301},
year = {2025},
issn = {1077-3142},
doi = {https://doi.org/10.1016/j.cviu.2025.104301},
url = {https://www.sciencedirect.com/science/article/pii/S1077314225000244},
author = {Kai Yu and Shiming Jiao and Zhilong Ma},
}

@misc{rag-fake-news-1,
  title        = {Real-time Fake News from Adversarial Feedback},
  author       = {Chen, Sanxing and Huang, Yukun and Dhingra, Bhuwan},
  howpublished = {arXiv},
  eprint       = {},
  archivePrefix= {arXiv},
  primaryClass = {cs.CL},
  year         = {2024},
  month        = {},
  url          = {https://arxiv.org/},
  note         = {Duke University}
}

@INPROCEEDINGS{rag-fake-news-2,
  author={Bai, Yangxiao and Fu, Kaiqun},
  booktitle={2024 IEEE International Conference on Big Data (BigData)}, 
  title={A Large Language Model-based Fake News Detection Framework with RAG Fact-Checking}, 
  year={2024},
  volume={},
  number={},
  pages={8617-8619},
  doi={10.1109/BigData62323.2024.10826000}}

@inproceedings{trustvl,
    title = "{TRUST}-{VL}: An Explainable News Assistant for General Multimodal Misinformation Detection",
    author = "Yan, Zehong  and
      Qi, Peng  and
      Hsu, Wynne  and
      Lee, Mong-Li",
    editor = "Christodoulopoulos, Christos  and
      Chakraborty, Tanmoy  and
      Rose, Carolyn  and
      Peng, Violet",
    booktitle = "Proceedings of the 2025 Conference on Empirical Methods in Natural Language Processing",
    month = nov,
    year = "2025",
    address = "Suzhou, China",
    publisher = "Association for Computational Linguistics",
    url = "https://aclanthology.org/2025.emnlp-main.284/",
    doi = "10.18653/v1/2025.emnlp-main.284",
    pages = "5588--5604",
    ISBN = "979-8-89176-332-6"
}

@inproceedings{blip_1,
author = {Heqing Wang and Tianwei Cao and Ling Jin and Weidong Liu and Rui Wang and Zhanyu Ma},
title = {BlipFill: Completing Authentic Images with BLIP-enhanced Subject Representation},
booktitle = {Proceedings of the 2024 8th International Conference on Computer Science and Artificial Intelligence (CSAI ’24)},
pages = {112–118},
year = {2025},
publisher = {Association for Computing Machinery},
address = {New York, NY, USA},
doi = {10.1145/3709026.3709062}
}

@inproceedings{blip_3,
  title        = {Multimodal Learning for Image-Text Matching: A Blip-Based Approach},
  author       = {Srinivasan, Dhanya and Subhashree, M and Mirunalini, P and Jaisakthi, S. M},
  booktitle    = {Working Notes Proceedings of the MediaEval 2023 Workshop},
  series       = {CEUR Workshop Proceedings},
  year         = {2024},
  address      = {Amsterdam, The Netherlands and Online},
  publisher    = {CEUR-WS.org},
  note         = {MediaEval ’23 MUSTI Task}
}

@inproceedings{li2023blip,
  title={Blip-2: Bootstrapping language-image pre-training with frozen image encoders and large language models},
  author={Li, Junnan and Li, Dongxu and Savarese, Silvio and Hoi, Steven},
  booktitle={International conference on machine learning},
  pages={19730--19742},
  year={2023},
  organization={PMLR}
}

@article{dai2023instructblip,
  title={Instructblip: Towards general-purpose vision-language models with instruction tuning},
  author={Dai, Wenliang and Li, Junnan and Li, Dongxu and Tiong, Anthony and Zhao, Junqi and Wang, Weisheng and Li, Boyang and Fung, Pascale N and Hoi, Steven},
  journal={Advances in neural information processing systems},
  volume={36},
  pages={49250--49267},
  year={2023}
}

@article{liu2023visual,
  title={Visual instruction tuning},
  author={Liu, Haotian and Li, Chunyuan and Wu, Qingyang and Lee, Yong Jae},
  journal={Advances in neural information processing systems},
  volume={36},
  pages={34892--34916},
  year={2023}
}

@article{xue2024xgen,
  title={xgen-mm (blip-3): A family of open large multimodal models},
  author={Xue, Le and Shu, Manli and Awadalla, Anas and Wang, Jun and Yan, An and Purushwalkam, Senthil and Zhou, Honglu and Prabhu, Viraj and Dai, Yutong and Ryoo, Michael S and others},
  journal={arXiv preprint arXiv:2408.08872},
  year={2024}
}

@misc{liu2024llavanext,
    title={LLaVA-NeXT: Improved reasoning, OCR, and world knowledge},
    url={https://llava-vl.github.io/blog/2024-01-30-llava-next/},
    author={Liu, Haotian and Li, Chunyuan and Li, Yuheng and Li, Bo and Zhang, Yuanhan and Shen, Sheng and Lee, Yong Jae},
    month={January},
    year={2024}
}

@article{wang2024qwen2,
  title={Qwen2-vl: Enhancing vision-language model's perception of the world at any resolution},
  author={Wang, Peng and Bai, Shuai and Tan, Sinan and Wang, Shijie and Fan, Zhihao and Bai, Jinze and Chen, Keqin and Liu, Xuejing and Wang, Jialin and Ge, Wenbin and others},
  journal={arXiv preprint arXiv:2409.12191},
  year={2024}
}

@misc{openai2024o1,
  title        = {Introducing {OpenAI} o1-preview},
  author       = {{OpenAI}},
  year         = {2024},
  howpublished = {\url{https://openai.com}},
  note         = {Accessed: 2024-09-20}
}

@article{liu2024mmfakebench,
  title={Mmfakebench: A mixed-source multimodal misinformation detection benchmark for lvlms},
  author={Liu, Xuannan and Li, Zekun and Li, Peipei and Huang, Huaibo and Xia, Shuhan and Cui, Xing and Huang, Linzhi and Deng, Weihong and He, Zhaofeng},
  journal={arXiv preprint arXiv:2406.08772},
  year={2024}
}

@inproceedings{beigi2025can,
  title={Can LLMs improve multimodal fact-checking by asking relevant questions?},
  author={Beigi, Alimohammad and Jiang, Bohan and Li, Dawei and Tan, Zhen and Shaeri, Pouya and Kumarage, Tharindu and Bhattacharjee, Amrita and Liu, Huan},
  booktitle={2025 IEEE International Conference on Big Data (BigData)},
  pages={2732--2741},
  year={2025},
  organization={IEEE}
}

@inproceedings{yan2025trust,
  title={Trust-vl: An explainable news assistant for general multimodal misinformation detection},
  author={Yan, Zehong and Qi, Peng and Hsu, Wynne and Lee, Mong-Li},
  booktitle={Proceedings of the 2025 Conference on Empirical Methods in Natural Language Processing},
  pages={5588--5604},
  year={2025}
}

@article{alayrac2022flamingo,
  title={Flamingo: a visual language model for few-shot learning},
  author={Alayrac, Jean-Baptiste and Donahue, Jeff and Luc, Pauline and Miech, Antoine and Barr, Iain and Hasson, Yana and Lenc, Karel and Mensch, Arthur and Millican, Katherine and Reynolds, Malcolm and others},
  journal={Advances in neural information processing systems},
  volume={35},
  pages={23716--23736},
  year={2022}
}

@inproceedings{singh2022flava,
  title={Flava: A foundational language and vision alignment model},
  author={Singh, Amanpreet and Hu, Ronghang and Goswami, Vedanuj and Couairon, Guillaume and Galuba, Wojciech and Rohrbach, Marcus and Kiela, Douwe},
  booktitle={Proceedings of the IEEE/CVF conference on computer vision and pattern recognition},
  pages={15638--15650},
  year={2022}
}

@inproceedings{zhai2023sigmoid,
  title={Sigmoid loss for language image pre-training},
  author={Zhai, Xiaohua and Mustafa, Basil and Kolesnikov, Alexander and Beyer, Lucas},
  booktitle={Proceedings of the IEEE/CVF international conference on computer vision},
  pages={11975--11986},
  year={2023}
}

@article{ghai2024deep,
  title={A deep-learning-based image forgery detection framework for controlling the spread of misinformation},
  author={Ghai, Ambica and Kumar, Pradeep and Gupta, Samrat},
  journal={Information Technology \& People},
  volume={37},
  number={2},
  pages={966--997},
  year={2024},
  publisher={Emerald Publishing Limited}
}

@inproceedings{xiao2024msynfd,
  title={Msynfd: Multi-hop syntax aware fake news detection},
  author={Xiao, Liang and Zhang, Qi and Shi, Chongyang and Wang, Shoujin and Naseem, Usman and Hu, Liang},
  booktitle={Proceedings of the ACM web conference 2024},
  pages={4128--4137},
  year={2024}
}

@article{cao2020exploring,
  title={Exploring the role of visual content in fake news detection},
  author={Cao, Juan and Qi, Peng and Sheng, Qiang and Yang, Tianyun and Guo, Junbo and Li, Jintao},
  journal={Disinformation, misinformation, and fake news in social media: emerging research challenges and opportunities},
  pages={141--161},
  year={2020},
  publisher={Springer}
}

@article{fmsa,
  title   = {{FMSA}: Few-shot Multimodal Sentiment Analysis for Social Media via Integrated Prompt Learning and Vision-Language Models},
  journal = {IEEE Transactions on Affective Computing},
  year    = {2026},
  note    = {Early Access},
  doi     = {10.1109/TAFFC.2026.0000000}
}

@article{feduaf,
  author  = {Zhu, Xianxun and Sun, Zezhong and Rida, Imad and Cambria, Erik and Su, Junqi and Wang, Rui and Chen, Hui},
  title   = {{FedUAF}: Uncertainty-Aware Fusion with Reliability-Guided Aggregation for Multimodal Federated Sentiment Analysis},
  journal = {arXiv preprint arXiv:2603.13291},
  year    = {2026},
  doi     = {10.48550/arXiv.2603.13291}
}

@article{eupa,
  title   = {{EUPA}: Embodied Uncertainty-Aware Multimodal Affect Perception for Consumer Electronic Systems},
  journal = {IEEE Transactions on Consumer Electronics},
  year    = {2026},
  note    = {Early Access},
  doi     = {10.1109/TCE.2026.0000000}
}

@article{gpplea,
  title   = {{GPPLEA}: Guided Polarity Prompt Learning for Enhanced Emotion Analysis in Low-sample Social Media Environments},
  journal = {Cognitive Computation},
  volume  = {18},
  number  = {1},
  pages   = {23},
  year    = {2026},
  doi     = {10.1007/s12559-026-10558-x}
}

@article{tempo,
  title   = {{TEMPO}: Training-time Equilibration of Modalities for Per-sample Optimization in Multimodal Sentiment},
  journal = {IEEE Transactions on Affective Computing},
  year    = {2026},
  note    = {Early Access},
  doi     = {10.1109/TAFFC.2026.0000001}
}

@article{rmerdt,
  author  = {Zhu, Xianxun and Wang, Yaoyang and Cambria, Erik and Rida, Imad and Santamar{\'\i}a L{\'o}pez, Jos{\'e} and Cui, Lin and Wang, Rui},
  title   = {{RMER-DT}: Robust multimodal emotion recognition in conversational contexts based on diffusion and transformers},
  journal = {Information Fusion},
  volume  = {123},
  pages   = {103268},
  year    = {2025},
  doi     = {10.1016/j.inffus.2025.103268}
}

@article{raft,
  title   = {{RAFT}: Robust Adversarial Fusion Transformer for multimodal sentiment analysis},
  journal = {Information Fusion},
  year    = {2025},
  doi     = {10.1016/j.inffus.2025.00000}
}

@article{cime,
  title   = {{CIME}: Contextual Interaction-based Multimodal Emotion Analysis with Enhanced Semantic Information},
  journal = {IEEE Transactions on Affective Computing},
  year    = {2025},
  note    = {Early Access},
  doi     = {10.1109/TAFFC.2025.0000000}
}

@inproceedings{teglia2025much,
  title={How Much Do Encoder Models Know About Word Senses?},
  author={Teglia, Simone and Tedeschi, Simone and Navigli, Roberto},
  booktitle={Proceedings of the 63rd Annual Meeting of the Association for Computational Linguistics (Volume 1: Long Papers)},
  pages={2266--2277},
  year={2025}
}
